\documentclass[review]{elsarticle}

\usepackage{amsmath}
\usepackage{amssymb}
\usepackage{mathpazo}
\usepackage{times}
\usepackage{graphicx,subfigure}
\usepackage{color}
\usepackage{multirow}
\newcommand{\mypm}{\mathbin{\mathpalette\@mypm\relax}}

\newcommand{\newchanged}[1]{\textcolor{black}{#1}}

\usepackage{lineno,hyperref}
\modulolinenumbers[5]

\journal{Neural Networks}

\begin{document}

\begin{frontmatter}

\title{Unsupervised Heart-rate Estimation in Wearables With Liquid States and A Probabilistic Readout}


\author[imecnl]{Anup Das\corref{mycorrespondingauthor}}
\cortext[mycorrespondingauthor]{Corresponding author}
\ead{anup.kumar.das@imec-nl.nl,akdas.nus@icloud.com}

\author[imecnl]{Paruthi Pradhapan}
\author[imecnl]{Willemijn Groenendaal}
\author[imecnl,tue]{Prathyusha Adiraju}
\author[imecnl]{\\Raj Thilak Rajan}
\author[imecbe,imecnl]{Francky Catthoor\corref{mycorrespondingauthor}}
\ead{Francky.Catthoor@imec.be}
\author[imecnl]{Siebren Schaafsma}
\author[uci]{\\Jeffrey L. Krichmar}
\author[uci]{Nikil Dutt}
\author[imecbe,imecnl]{Chris Van Hoof}

\address[imecnl]{Stichting IMEC Nederland, High Tech Campus 31, Eindhoven 5656 AE, The Netherlands}
\address[tue]{Eindhoven University of Technology, De Zaale, Eindhoven 5612 AZ, The Netherlands}
\address[imecbe]{IMEC Leuven, Kapeldreef 75, 3001 Heverlee, Belgium}
\address[uci]{University of California, Irvine, Irvine, CA 92697-5100, USA}

\begin{abstract}
{Heart-rate estimation is a fundamental feature of modern wearable devices. In this paper we propose a machine intelligent approach for heart-rate estimation from electrocardiogram (ECG) data collected using wearable devices. The novelty of our approach lies in (1) encoding spatio-temporal properties of ECG signals directly into spike train and using this to excite recurrently connected spiking neurons in a Liquid State Machine computation model; (2) a novel learning algorithm; and (3) an intelligently designed unsupervised readout based on Fuzzy c-Means clustering of spike responses from a subset of neurons (Liquid states), selected using particle swarm optimization. Our approach differs from existing works by learning directly from ECG signals (allowing personalization), without requiring costly data annotations. Additionally, our approach can be easily implemented on state-of-the-art spiking-based neuromorphic systems, offering high accuracy, yet significantly low energy footprint, leading to an extended battery life of wearable devices. We validated our approach with CARLsim, a GPU accelerated spiking neural network simulator modeling Izhikevich spiking neurons with Spike Timing Dependent Plasticity (STDP) and homeostatic scaling. A range of subjects are considered from in-house clinical trials and public ECG databases. Results show high accuracy and low energy footprint in heart-rate estimation across subjects with and without cardiac irregularities, signifying the strong potential of this approach to be integrated in future wearable devices.}
\end{abstract}


\end{frontmatter}


\section{Introduction}
\begin{figure}[t]
	\begin{center}
		\includegraphics[width=4.5in]{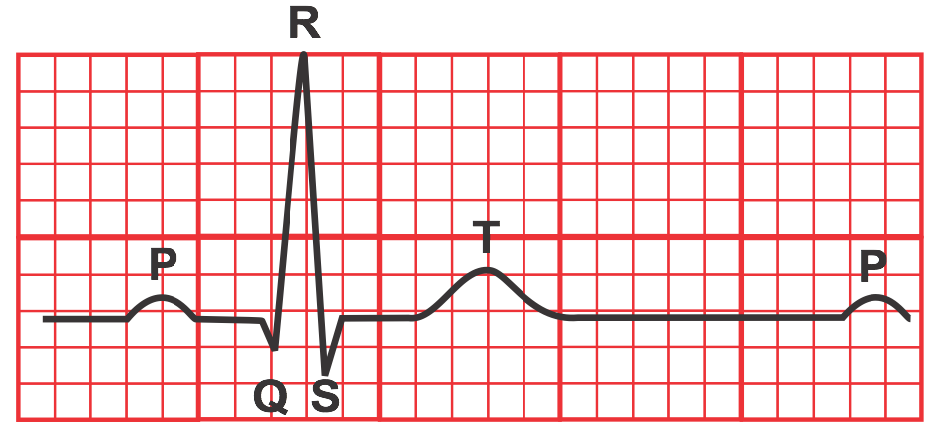}
	\end{center}
	\caption{QRS complex identification from ECG signals.}
	\label{fig:qrs}
\end{figure}

Heart-rate monitoring is ubiquitous in modern wearable devices such as a smart watch \citep{phan2015smartwatch,aarts2006apparatus} or Electrocardiogram (ECG) necklace \citep{penders2011low}. ECG sensors \citep{mukhopadhyay2015wearable,gyselinckx2005human} attached to these devices monitor the electrical potential caused by the systolic activity of heart and then propagated through cardiac muscles. The recorded electrical data are post-processed, either locally on the sensor \citep{van2015345} or on a device \citep{7367297} attached to the sensor to estimate heart-rate. QRS pattern identification (see Figure \ref{fig:qrs}) from ECG is fundamental to heart-rate estimation. Although QRS detection has achieved significant maturity over time \citep{kohler2002principles}, recent advances in wearable healthcare \citep{gyselinckx2005human,otto2006system} have motivated researchers to revisit QRS. This is due to 

\begin{itemize}
	\item ECG readings from wearable sensors are contaminated with motion artifacts and baseline drifts; and
	\item devices integrating wearable sensors are constrained in terms of area, power consumption and computational capabilities.
\end{itemize}

Several approaches have been proposed recently to detect QRS complex from wearable ECG using software techniques. A level crossing approach is proposed in \citep{ravanshad2014level}, where a modified level-crossing analog-to-digital converter is introduced to convert analog ECG data to a meaningful representation. Based on this representation, an algorithm is proposed to detect the RR intervals. A time-frequency representation (called the S-Transform) is introduced in \citep{zidelmal2014qrs} to isolate QRS complexes in time-frequency domain. Shanon energy is computed on these isolated spectrums to localize R-waves in time domain. A real-time signal processing approach is proposed in \citep{Karimipour2014153}, which includes high frequency noise filtering and baseline drift reduction using discrete wavelet transform. In \citep{RKSPL2014}, a threshold independent approach is proposed based on first derivative of ECG signals. There are also other deterministic approaches for QRS detection. For a summary of these approaches, readers are referred to \citep{Jain2017362}. Some of the recent works have also addressed low power hardware implementations \citep{van2015345,deepu2016ecg,ieong2016sub,kim2017energy}. {Software-based QRS detection techniques require general purpose computing platforms (such as a CPU core). Power consumption of these techniques is usually of the order of $\approx10\mu$W. One advantage of these approaches is the ease of updating an existing algorithm or implementing a new use-case (such as arrhythmia detection), with software updates only (no hardware change is required). Hardware-based QRS detection techniques require dedicated hardware to implement the chosen algorithm. Although, sub-$\mu$W power consumption is achieved in hardware-based techniques, limited flexibility is offered by these designs to implement a new algorithm or a new use-case. For these techniques, the usual approach is to do feature extraction or QRS detection on the sensor with use-cases such as arrhythmia detection implemented on the device \citep{tekeste2017nano}.}

{In recent years, artificial intelligent approaches have been investigated for QRS detection to address flexibility, while consuming lower power, an important consideration for power-constrained wearable devices.} Artificial neural network has been used to carry out the classification task in \citep{silpioTSP98}. Such a network forms the basis for ECG analysis including arrhythmia, myocardial ischemia and other chronic alterations. K-nearest neighbor classifier is used for QRS complex detection from ECG in \citep{Saini2013331}. The ECG signal is post-processed using a digital band-pass filter to reduce false detection with gradient of the signal used as a feature for the classifier. Radial basis function is used for QRS complex identification in \citep{Arbateni2014438}. Similar to the previous approach, this technique uses filters as a pre-processing step with a centering approach for adjusting the R-peak positions. Support vector machine is used in \cite{magrans2016myocardial} to classify QRS segments. An 11 layer deep convolution neural network based approach is proposed in \citep{acharya2017automated,acharya2017automated2} to classify ECG segments. All these approaches use the classical neural network approaches with supervised learning. {Success of these approaches depends on availability of large amount of hand labeled data, generic enough to be applied to a broad range of subjects with and without cardiac irregularities.} 

In this work we use spiking neural networks \citep{maass1997networks} for heart-rate estimation. These are powerful and biologically realistic computation models, inspired by the dynamics of human brain. Spiking neural networks are gaining popularity in solving complex pattern recognition \citep{buonomano1999neural,kasabov2013dynamic}, function approximation \citep{iannella2001spiking} and image classification \citep{diehl2015fast,diehl2015unsupervised,samadi2017deep} tasks. Another reason for widespread success of spiking neural networks is their efficient VLSI implementation \citep{indiveri2006vlsi,fusi1998collective,hsieh2012vlsi}. Examples are the large scale neuromorphic computing\footnote{The term neuromorphic computing was first coined in \citep{mead1990neuromorphic}.} systems such as TrueNorth \citep{akopyan2015truenorth}, CxQuad \citep{indiveri2015neuromorphic}, NeuCube \citep{kasabov2014neucube}, SpiNNaker \citep{khan2008spinnaker}, NeuroGrid \citep{benjamin2014neurogrid} and HICANN \citep{schemmel2010wafer}, among others. {Some of these systems are originally designed for high-performance computing (e.g., TrueNorth and SpiNNaker), while others are designed for low-power embedded systems (e.g., CxQuad and HICANN). Readers are referred to \cite{schuman2017survey} for a recent survey of neuromorphic hardware. In this work we report energy usage on CxQuad, a low power spiking hardware with 1024 neurons and 64K synapses (\citep{indiveri2015neuromorphic}).}

{Our work differentiates from existing studies on ECG-based heart-rate estimation by (1) using spiking neural networks, which can be implemented on energy efficient neuromorphic hardware; (2) encoding analog ECG signal directly into spike trains, which are then used to excite the network of spiking neurons; and (3) designing an unsupervised readout, facilitating learning from subject-specific ECG to estimate heart-rate. The overall approach allows personalization and eliminates the need to manually annotate training data. We envision our approach to be integrated inside an ECG sensor node. Analog ECG signal is encoded directly into spikes, which are used to excite a reservoir of recurrently connected spiking neurons. This is inspired by the computation model of Liquid State Machine  \citep{maass2002real}. Neurons in the architecture are interconnected using plastic synapses, with weight updates using Spike Timing Dependent Plasticity (STDP) \citep{rao2001spike,brader2007learning}, an important form of Hebbian Learning. Additionally, homeostatic synaptic scaling \citep{liu2011learning,galtier2013biological,carlson2013biologically} is used to stabilize the plastic mechanism, preventing run-away behaviors. At the readout stage of the architecture, we use particle swarm optimization \citep{eberhart1995new} to select contributions from a subset (called the winning set) of the spiking neurons. Cumulative responses from these winning neurons are clustered to infer heart-rate using Fuzzy c-Means clustering \citep{bezdek1984fcm}. To validate our approach we used CARLsim \citep{beyeler2015carlsim} spiking neural network simulator with ECG data from in-house clinical trials and also open-source databases. We compared our approach with (1) deterministic QRS detection technique of \citep{van2015345}; (2) the neural network based technique of \citep{acharya2017automated}; and the support vector machine based technique of \citep{magrans2016myocardial}\footnote{Original flavors of \citep{acharya2017automated,magrans2016myocardial} are adapted for QRS detection and heart-rate estimation only.}. Results demonstrate high accuracy of our approach, with applicability to subjects with and without cardiac irregularities.}

\noindent\textbf{Contributions: }{Following are our novel contributions:}
\begin{itemize}
	\item {a technique to encode time-series data to spike train, capturing its spatio-temporal characteristics;}
	\item {a novel architecture inspired by the computation model of Liquid State Machine;}
	\item {a novel learning-rule with soft winner-take-all implementation, facilitating temporal pattern detection;}
	\item {an unsupervised readout for heart-rate estimation using Fuzzy c-Means;}
	\item {a technique to improve clustering accuracy by selecting neurons responses using swarm intelligence; and}
	\item {a real-life medical benchmark for neuromorphic computing community.}
\end{itemize}

\section{Overview of our Approach}
{Our approach is based on the computational model of Liquid State Machine (LSM) \citep{maass2002real}, which consists of a pre-processing unit (called the ``Liquid") with recurrently connected spiking neurons and a readout unit to decode/interpret the internal states of the liquid. LSM can implement any non-linear transformation of the input. Internally, the Liquid integrates temporal information into linearly independent Liquid states, without computing precisely any nonlinear transformation; the readout has no notion of temporal aspects and learns to map Liquid states to the function approximation/classification. Following are some characteristics of LSM that are relevant for the heart-rate estimation problem we aim to solve in this work.}

\begin{figure}[t]
	\hfill
	\begin{center}
		\includegraphics[width=5in]{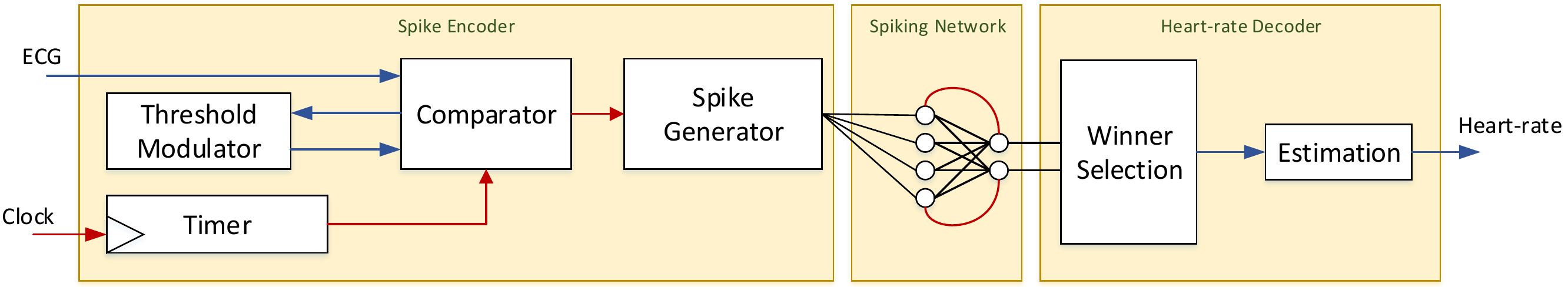}
	\end{center}
	\caption{Components of our approach based on Liquid State Machine (LSM). There are three main units: Spike Encode, Spiking Neural Network and Heart-Rate Decoder.}
	\label{fig:arch}
\end{figure}

\begin{itemize}
	\item {Recurrently connected neurons in the Liquid can implement memory of the input spikes at many (potentially infinitely many) preceding time instances. This is required as morphological and certain other dynamic features of the ECG are a result of cardiac changes over a period of time, which needs to be incorporated for personalized heart-rate estimation (temporal).}
	\item {Separating the computation into Liquid and readout allows to instantiate multiple readouts (implementing a large range of functionalities) using the same Liquid. In this work we discuss one such readout instantiation using unsupervised learning algorithm, directed by the statistics of the spike input to the Liquid. Additionally in the context of wearables, the Liquid can be instantiated inside the sensor with multiple readouts implemented on the device (flexibility).}
	\item {LSM allows real-time computation with fast and robust learning from streaming data. This makes the computational model specially suitable for heart-rate estimation, where on-sensor subject-specific learning is desirable.}
\end{itemize}


{An overview of our approach is shown in Figure \ref{fig:arch}. Spikes generated from the encoder (\emph{Spike Encoder}) are used to excite the Liquid (\emph{Spiking Network}). Liquid states are used to infer heart-rate in the readout unit (\emph{Heart-Rate Decoder}). We envision the spike encoder and the spiking network (Liquid) to be integrated inside the sensor, while the readout resides on the wearable device. In this way, spatio-temporal characteristics of the ECG processed by the Liquid can be used to implement multiple use-cases (as separate readouts) on a wearable device, depending on user preference and market needs. Examples include arrhythmia detection or ECG-based authentication.}

\subsection{Spike Encoder}
\label{sec:spike_encoder}
Information in a spiking neural network can be encoded using two techniques -- rate-based coding and temporal coding \citep{van2001rate}. Rate-based coding encodes information as number of spikes within an encoding window without considering the temporal characteristics of the signal. Rate-based encoding has been successfully applied to spatial classification tasks such as handwritten digit recognition \citep{diehl2015unsupervised}. Temporal coding \citep{van2001novel}, on the other hand, encodes information as inter-spike intervals, capturing the spatio-temopral structure of the input signal. Temporal encoding has been successfully applied for time-series processing such as speech processing \citep{Tavanaei2017191} and EEG-based brain-machine interface \citep{CorradiTBME}. For heart-rate estimation using ECG, temporal characteristics around QRS complexes need to encoded as inter-spike intervals and therefore, temporal coding is adopted in this work.

\begin{figure}[t]
	\hfill
	\begin{center}
		\includegraphics[width=2.5in]{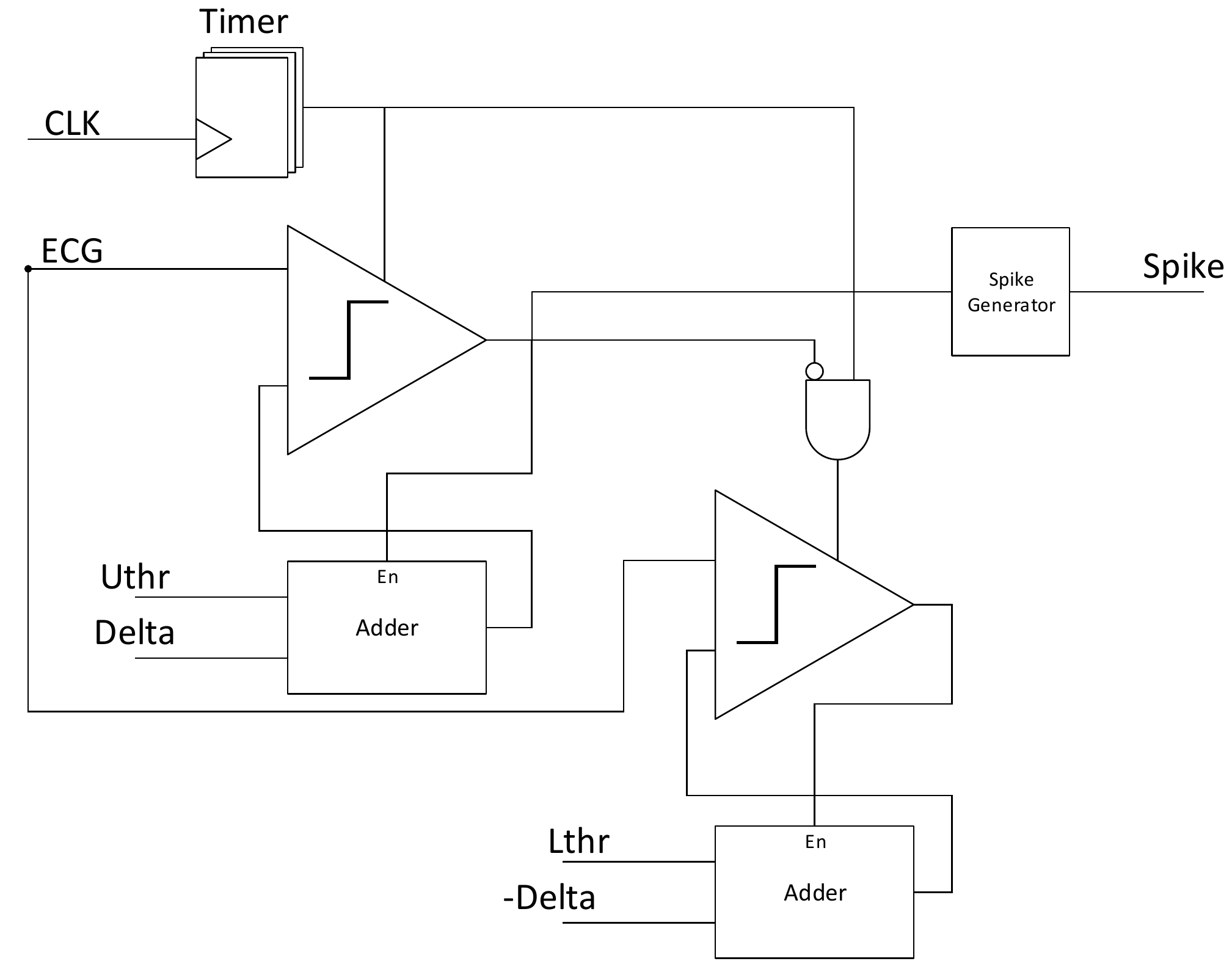}
	\end{center}
	\caption{Circuit for spike generation from ECG.}
	\label{fig:circuit}
\end{figure}

\subsubsection{Overview of the Spike Encoding Circuit Design}
\begin{figure}[t]
	\hfill
	\begin{center}
		\includegraphics[width=4.5in]{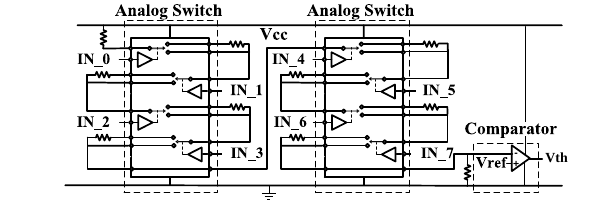}
	\end{center}
	\caption{Voltage comparator from author's earlier work \citep{balsamo2016hibernus}.}
	\label{fig:voltage_comparator}
\end{figure}

{The spike encoder encodes input ECG signal to inter-spike intervals using a combination of threshold modulators, voltage comparator, spike generator and a timer. This is shown in Figure \ref{fig:circuit}. The threshold modulators are implemented using full adders. Such adders can be efficiently implemented using thin film transistors \citep{bahubalindruni2016novel}. Reference analog voltage comparator and spike generation implementations are discussed in \citep{li2014ultra,du2011bio}. The timer in Figure \ref{fig:circuit} can be implemented using D flip flops (DFF), clocked using an external crystal oscillator. The DFF implements clock division to generate the trigger intervals for the comparators. In this work, we use 2 ms trigger intervals, i.e. the spike encoder circuit operates at 500 Hz. Accuracy-power consumption trade-off obtained by varying this trigger interval is left as future work. It is to be noted that the comparator circuit can be shared to reduce power consumption. Figure~\ref{fig:voltage_comparator} shows the schematic of such an arrangement for a sensor node~\citep{balsamo2016hibernus} with a maximum propagation delay of $5\mu S$. The two analog switches multiplex the signal propagated to the comparator\footnote{Detailed circuit analysis of the spike encoder is beyond the scope of this work.}. In the next section we describe the high level working of the spike encoding circuit.}

\begin{figure}[t]
	\hfill
	\begin{center}
		\includegraphics[width=4.5in]{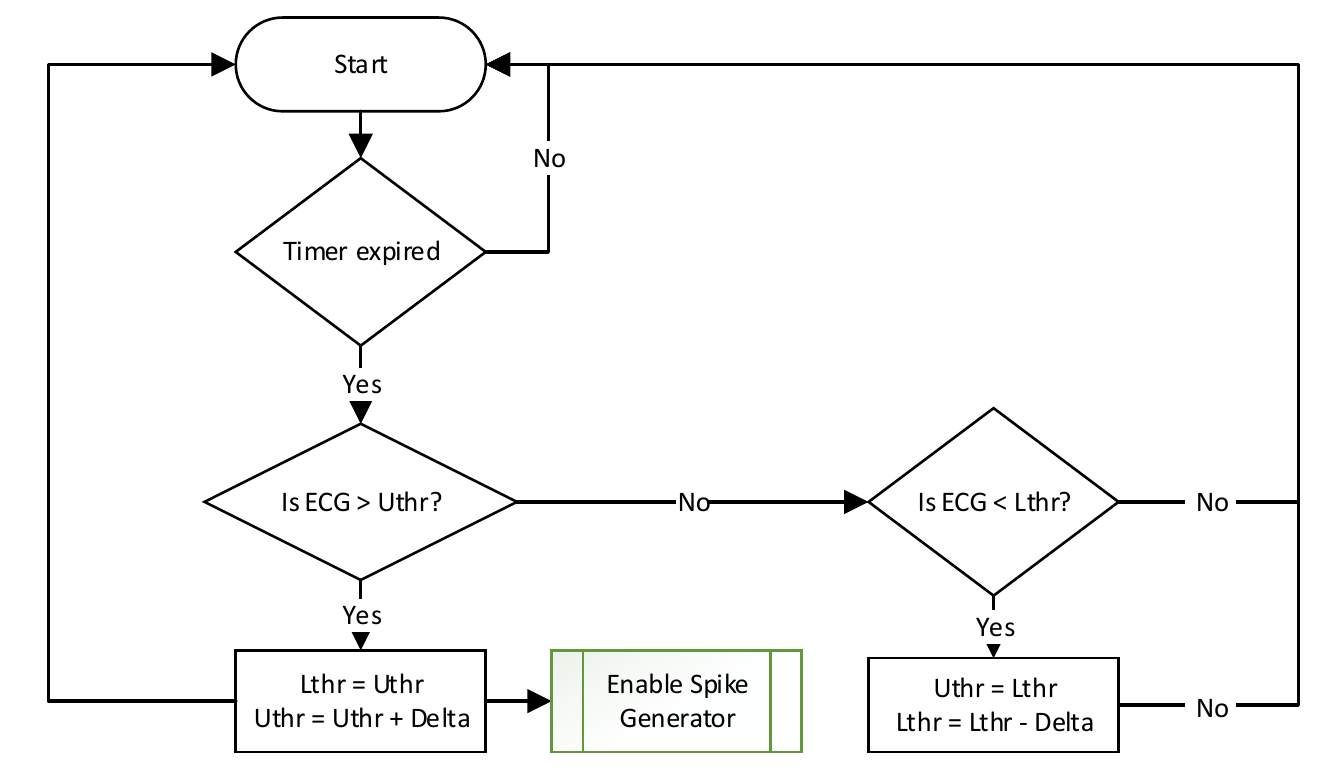}
	\end{center}
	\caption{Flowchart for spike encoding.}
	\label{fig:flow}
\end{figure}

\begin{figure}
	\centering     
	\subfigure[Spike generation for a typical ECG signal.]{\label{fig:ECG_input}	\includegraphics[width=4.5in]{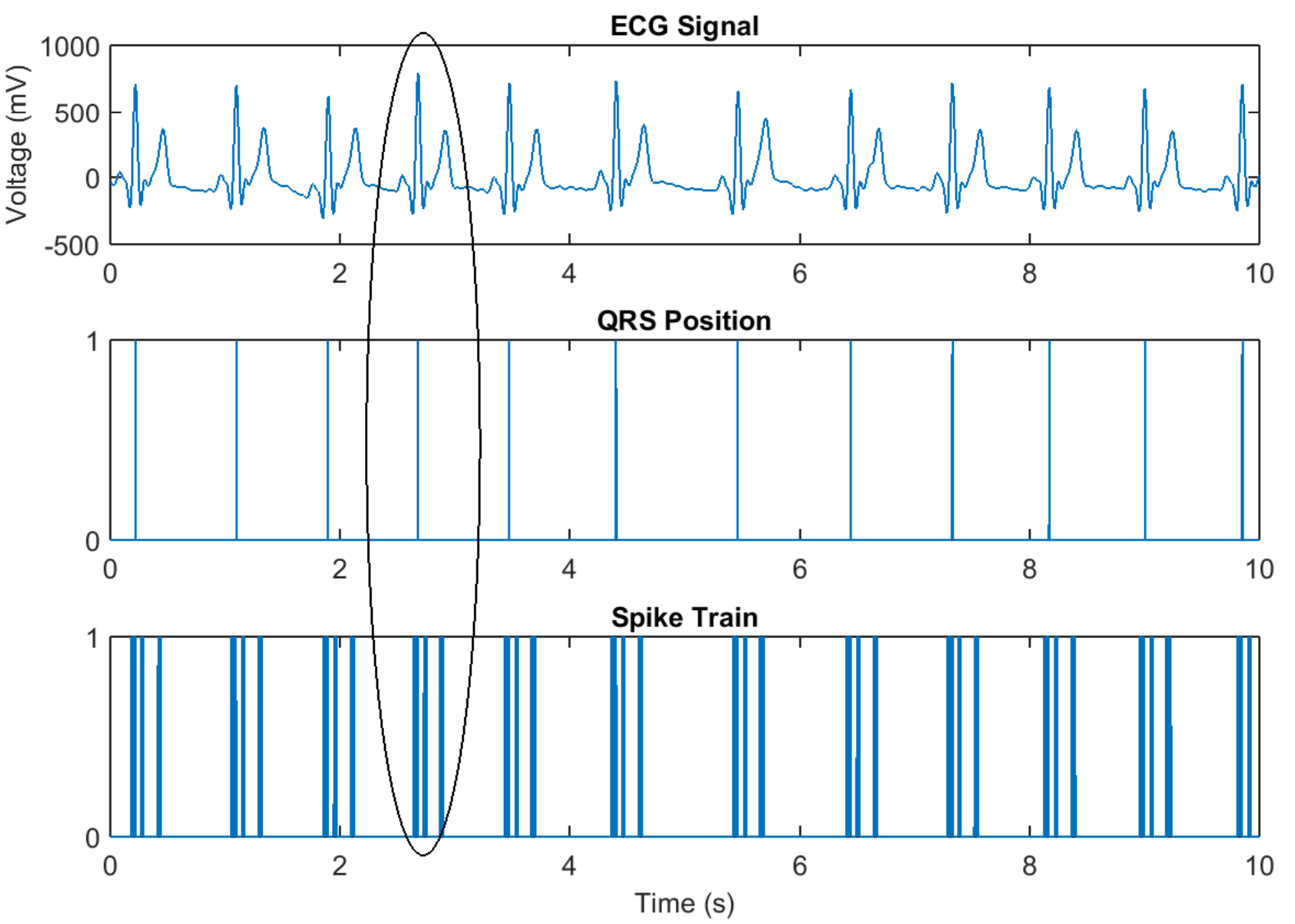}}
	\subfigure[Spike generation for a zoomed section of the ECG signal.]{\label{fig:any_input}	\includegraphics[width=4.5in]{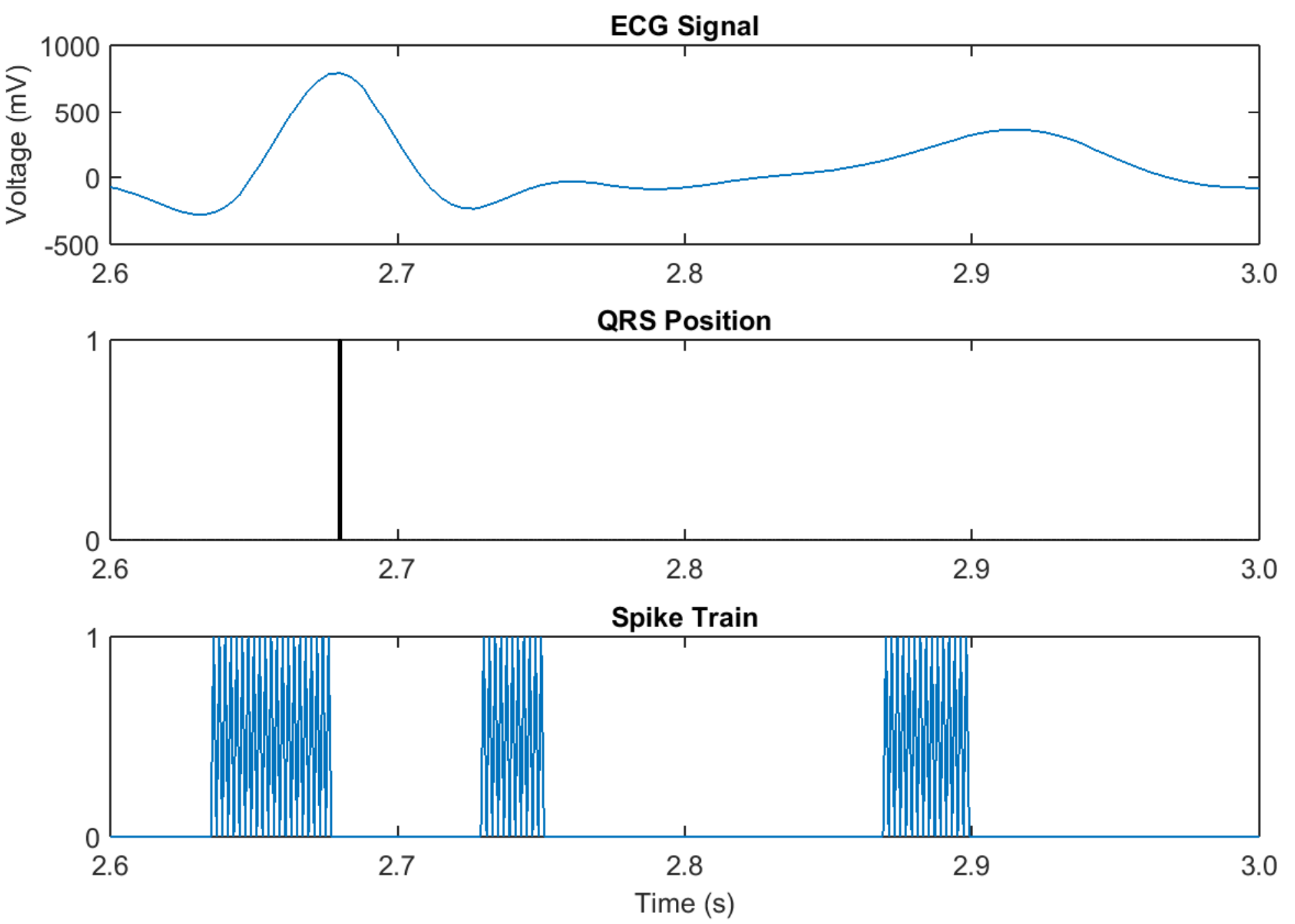}}
	\caption{Response of the spike encoder to input ECG signals. For Figure (1), The top sub-figure shows a segment of the ECG signal. The middle sub-figure shows the QRS peaks, which are our regions of interest. Finally, the bottom sub-figure shows the spikes generated for the ECG signal of the top sub-figure.}
\end{figure}

\subsubsection{Working Principle of the Spike Encoder}
The working principle of the spike encoder is described considering three scenarios -- ECG signal is rising, ECG signal is falling and ECG signal is stable within a time window \texttt{Delta}. The algorithm for the spike encoder is shown as a flowchart in Figure \ref{fig:flow}. The encoding is based on two thresholds -- \texttt{Lthr} and \texttt{Uthr} ($\texttt{Uthr} > \texttt{Lthr}$). Voltage comparisons are performed at a fixed intervals controlled using the timer. At every interval, the comparator is enabled to compare the ECG voltage and the threshold \texttt{Uthr}. For simplicity, we assume that the ECG signal is rising. The voltage comparison is positive, triggering the following sequence of events.
\begin{itemize}
	\item threshold update: \texttt{Lthr} = \texttt{Uthr} and \texttt{Uthr} = \texttt{Uthr} + \texttt{Delta}
	\item enable spike generator to emit one spike
	\item restart timer and return to wait
\end{itemize}

If the result of the comparison is negative (meaning the ECG signal is either falling or stable within \texttt{Delta}), a second comparison is performed, where the ECG voltage is compared with the threshold \texttt{Lthr}. If this comparison is positive (in the case when the ECG signal is falling), the following sequence of events are triggered.
\begin{itemize}
	\item threshold update: \texttt{Uthr} = \texttt{Lthr} and \texttt{Lthr} = \texttt{Lthr} - \texttt{Delta}
	\item restart timer and return to wait
\end{itemize}

Finally, if the result of the second comparison is also negative (in the case when the ECG signal is stable within \texttt{Delta}), thresholds are not updated and timer is restarted. Following are the specific characteristics of this approach.
\begin{itemize}
	\item Threshold updates are performed to track the ECG signal in the upward or downward directions. 
	\item No spikes are generated when the ECG signal is falling. This is a design choice we use in this work as the spatio-temporal characteristics of the QRS can be captured using the rising part of the voltage waveform.
	\item No spikes are generated if the ECG signal is stable.
\end{itemize}


Figure \ref{fig:ECG_input} shows the spike generation from a segment of the ECG signal. Also highlighted in this figure are the regions of interest i.e., the QRS peaks (shown in the middle figure). As discussed before and this can be seen quite well from this figure, spike temporal coding captures important spatio-temporal characteristics from the input signal in the form of inter-spike intervals. Figure \ref{fig:any_input} shows a zoomed part of the ECG signal. As discussed before, we use spikes generated from the rising signal only to reduce data transfer without compromising accuracy.



\subsection{Spiking Neurons-Based Liquid}
{This section describes the Liquid, starting with the neuron and synapse model and followed by the Liquid topology. The readout unit is described in Section \ref{sec:readout}.}

\subsubsection{Neuron and Synapse Model}
The Liquid in our approach is built using the Izhikevich neuron \citep{izhikevich2003simple} model with the following two-dimensional ordinary differential equations
\begin{align}
\label{eq:eq1}
\frac{dv}{dt} = 0.04v^2 + 5v + 140 - u + I_{\text{syn}} \text{ ,~~for the membrane potential~} v\nonumber \\
\frac{du}{dt} = a(bv - u) \text{ ,~~for the recovery variable~} u 
\end{align}

\begin{equation}
\label{eq:eq1a}
\text{if } v > 30\text{mV, then } \begin{cases}
v = c\\
u = u + d
\end{cases} 
\end{equation}
where $I_{\text{syn}}$ is the synaptic current to the neuron. In contrast to simple models such as leaky integrate-and-fire type neurons, the Izhikevich neuron is more biologically realistic with voltage reset occurring at the peak of the spike (as opposed to occurring at threshold) and an instantaneous reset of the membrane potential when the membrane voltage reaches a cutoff voltage (Equation \ref{eq:eq1a}). Additionally, Izhikevich neurons can be efficiently implemented in hardware \citep{demirkol2011low}. \newchanged{Table \ref{tab:param} reports the parameters used for simulation. The subscript \texttt{E} is used to indicate excitatory neurons i.e., neurons that are glutamatergic, while the subscript \texttt{I} is used for inhibitory type neurons, i.e., neurons with GABAergic synapses. These parameters are chosen to model excitatory neurons as regular spiking neurons and inhibitory neurons as fast spiking neurons \cite{izhikevich2003simple}.}

\begin{table}[t]
	\renewcommand{\arraystretch}{0.8}
	\caption{Spiking neural network parameters used for simulation. Neuron parameters are that for Izhikevich neuron \citep{izhikevich2003simple}. Remaining parameters (initial weights/E-STDP/I-STDP/homeostasis) are a set of values during a specific iteration (selected randomly) of the design space exploration process.}
	\label{tab:param}
		\begin{center}
			\begin{tabular}{|c|c|c|}\hline
				Class & Parameters & Value\\
				\hline
				\multirow{4}{*}{Neuron} & $a_E$, $a_I$ & 0.02, 0.1\\
				 & $b_E$, $b_I$ & 0.2, 0.2\\
				 & $c_E$, $c_I$ & -65, -65\\
				 & $d_E$, $d_I$ & 8, 2\\
				\hline
				Initial synaptic strengths & $W_0$  & $0.1\pm 0.05$\\
				\hline
				\multirow{4}{*}{E-STDP/I-STDP} & $A_{+_E}$, $A_{+_I}$ & 0.1, -0.1\\
				 & $\tau_{+_E}$, $\tau_{+_I}$ & 20, 20\\
				 & $A_{+_E}$, $A_{+_I}$ & -0.1, 0.1\\
				 & $\tau_{-_E}$, $\tau_{-_I}$ & 20, 20\\
				 \hline
				 Homeostasis (Excitatory) & $\alpha,T,R_{\text{target}}$ & 0.1, 10, 35\\
				 Homeostasis (Inhibitory) & $\alpha,T,R_{\text{target}}$ & 0.1, 2, 3.5\\
				\hline
			\end{tabular}
		\end{center}
\end{table}

Current in the post-synaptic neuron $j$ is computed as
\begin{equation}
\label{eq:eq2}
I_j = \sum_{i=0}^{n_j} i_{ij}\cdot w_{ij}
\end{equation}
where $n_j$ is the total number of pre-synaptic neurons connected to the $j^{\text{th}}$ post-synaptic neuron, $i_{ij}$ is the current through the synaptic connection between $i^{\text{th}}$ pre-synaptic neuron and $j^{\text{th}}$ post-synaptic neuron and $w_{ij}$ is the corresponding weight. The synaptic currents $i_{ij}$ are modeled using conductance changes and therefore, decay over time.

\subsubsection{Liquid Topology}
\noindent \textbf{Connection Probability and Synaptic Delay:} Our Liquid topology consists of three layers -- input, recurrent and output. The first layer is the input layer, which generates spikes (encoded directly from the input ECG). The second layer is the recurrent layer and consists of $N = N_E + N_I$ recurrently connected neurons, where $N_E$ is the number of excitatory neurons and $N_I$ is the number of inhibitory neurons. Motivated by the anatomy of a mammalian cortex \citep{abeles1991corticonics,bock2011network}, our framework consists of $N_I = 0.25\cdot N_E$. Neurons in the second layer are interconnected satisfying the following -- excitatory neurons of the second layer can connect to any neurons (excitatory as well as inhibitory neurons), but the inhibitory neurons are only connected to other excitatory neurons. This decision is influenced by lack of evidence of recurrence in inhibitory neural assembles in mammalian cortex \citep{abeles1991corticonics}. Connection probabilities among the neurons are as follows: excitatory to excitatory neurons have connection probabilities of 0.01; excitatory to inhibitory and inhibitory to excitatory neurons have connection probabilities of 0.1. These probabilities are similar to that used in \citep{pfeil2013six}. \newchanged{Initial synaptic strengths $W_0$ are reported in Table \ref{tab:param}. Changes in synaptic strength are bounded between 0 and $10\times W_0$. Synaptic connection delays are selected randomly between 1ms and 2ms.}
	
\noindent \textbf{Connection Pruning implementing Soft Winner-Take-All:} Inhibitory to excitatory connections are pruned such that an inhibitory neuron is never connected to the excitatory neuron from which it receives a connection. This creates lateral inhibition in the Liquid topology. As an additional step, we fine tuned excitatory and inhibitory synaptic conductances, such that the lateral inhibition is neither too weak (which means lateral inhibition has no influence to distinguish QRS complexes), nor too strong (which means once a winner is selected, the winner prevents other neurons from firing). This implements a soft version of the winner-take-all strategy.

\noindent \textbf{Output Connections:} {The third layer in the LSM Liquid consists of neurons which are interpreted in the readout unit. In most LSM architectures, the output neurons reside inside the readout.} Neurons in the input layer are connected to the excitatory neurons of the second layer using plastic synapses, which are adjusted using inter-spike intervals (discussed in the next section). Learning of connections from the recurrent layer to the output layer is discussed in Section \ref{sec:readout} as part of the readout unit.

\subsubsection{Learning using STDP and Homeostatic Plasticity}
{In our framework, synaptic weight updates are disabled after a time interval interval $T_i$. The time interval 0 (start of ECG) to $T_i$ is training phase of the spiking neural network. In this phase, weights are updated using spike timing dependent plasticity (STDP) \citep{rao2001spike,brader2007learning}. STDP uses correlation between the spikes (based on spike times) to derive a potential causal (or anti-causal) relationship between the pre- and post-synaptic neurons. This correlation is then used to influence the weight changes \citep{sjostrom2010spike}\footnote{This is contrary to supervised learning such as using error backpropagation \citep{bohte2002error}, where a teacher signal is used to adjust the weights.}. Synaptic weights connecting an excitatory neuron to any other neurons are updated using excitatory-type STDP (E-STDP), while synaptic weights connecting an inhibitory neuron to an excitatory neurons are updated using inhibitory-type STDP (I-STDP). In this work, STDP (E-STDP/I-STDP) is realized using an exponential function}
\begin{equation}
\label{eq:eq2a}
\Delta W = \begin{cases}
A_+ \texttt{exp}(\frac{-\Delta t}{\tau_+}) & \text{for } \Delta t > 0\\
A_- \texttt{exp}(\frac{-\Delta t}{\tau_-}) & \text{otherwise}
\end{cases}
\end{equation}


To prevent run-away behavior, homeostatic synaptic scaling \citep{carlson2013biologically} is used. The combined weight change equation is given by
\begin{equation}
\label{eq:eq3}
\frac{dw_{ij}}{dt} = \Bigg[\alpha\cdot w_{ij}\Big(1-\frac{R_{avg}}{R_{target}}\Big) + \beta(\Delta w_{\text{before}_{ij}} + \Delta w_{\text{after}_{ij}})\Bigg]\cdot K
\end{equation}
where $\alpha$ is the homeostatic scaling factor, $\beta$ is the STDP scaling factor, $R_{avg}$ is the average firing rate of a neuron over a large period of time, $R_{target}$ is the predefined firing rate of the neuron (a design choice), $\Delta w_{\text{before}_{ij}}$ (and $\Delta w_{\text{after}_{ij}}$) is the pre-before-post (and pre-after-post) weight change contributions and $K$ is the scalability factor
\begin{equation}
\label{eq:eq4}
K = \frac{R_{avg}}{T\Big(1+|1-\frac{R_{avg}}{R_{target}}|\cdot\gamma\Big)}
\end{equation}
where $T$ is the time duration over which the firing rate is being averaged and $\gamma$ is a constant factor. Our exploration involves tuning the parameters of these equations.

\subsubsection{Implementation Framework of Spiking Neural Network}
\newchanged{The spiking neural network (spiking neurons and learning rule) is implemented using CARLsim \citep{beyeler2015carlsim}, a GPU-accelerated library for simulating spiking neural network models with a high degree of biological detail. We have performed exploration with different number of excitatory and inhibitory neurons. Based on these explorations, the spiking network in our approach is built with 64 excitatory and 16 inhibitory neurons.}

\subsection{Probabilistic Heart-rate Decoder as LSM Readout}
\label{sec:readout}
{The heart-rate decoder of Figure \ref{fig:arch} is the readout stage of our implementation of LSM computation model. Novelty of this readout is its ability to personalize to a subject (with and without cardiac irregularities) in an unsupervised approach, learning directly from the ECG signal processed using the Liquid. For this purpose we use Fuzzy c-Means clustering on a subset of output neurons (called the winning set). Cluster center selection and winning neurons selection sub-problems are jointly solved using our proposed particle swarm optimization technique. This optimization problem is solved once during initialization; subsequently, the readout evaluates cluster membership only using the evaluated centers\footnote{This approach will be extended in future to periodically carry out the initialization (optimization), to take into account major shift in ECG pattern during lifetime of a device or multiple user of the same device.}. Finally, we compute discrete heart-rate probability from cluster membership using Poisson Binomial Distribution, which is then used to calculate the expected heart-rate.}

We start our formulation by defining two intervals -- the \texttt{spike integrate interval} (\textbf{SI}) and the \texttt{heart-rate inference interval} (\textbf{HI}). Spike responses from a neuron are integrated (i.e., counted) in every \textbf{SI} interval. In the remainder of this work, the term \emph{spike response} is used to denote the spike count of a neuron in a \textbf{SI} interval, which is the smallest unit of time for our purpose. In our experimental setup, \textbf{SI} = 100 ms and \textbf{HI} = 1 min, allowing the heart-rate estimated in a \textbf{HI} interval to be directly interpreted as beats per minute (bpm). Each \textbf{HI} interval is composed of 600 \textbf{SI} intervals. Spike responses from the neurons in a \textbf{HI} interval are clustered into two classes (QRS and no-QRS) using the Fuzzy c-Mean algorithm \citep{bezdek1984fcm}, where cluster membership of a response is given by the probability distribution over the cluster. We define probability of success for a neuron response as the probability that the neuron response belong to the QRS cluster. Using the Fuzzy c-Mean outcomes as Bernoulli trails each with its characteristics probability of success, we compute the discrete probability distribution, which describes a fixed number of successes in the \textbf{HI} interval. This distribution is then used to compute the expected heart-rate for the \textbf{HI} interval. Following sections formally describe these steps.

\subsubsection{Fuzzy c-Means Clustering of Spike Count}
The Fuzzy c-Means (FCM) \citep{bezdek1984fcm} is a probabilistic clustering algorithm, where instead of computing the absolute membership of an item belonging to a cluster (hard decisions as in K Means clustering \citep{macqueen1967some}), FCM computes the likelihood of an item belonging to a cluster (soft or fuzzy  decisions).

We define the following notations to ease the problem formulation.
\begin{align*}
n &= \text{number of output neurons}\\
n_s &= \text{number of \textbf{SI} time segments in a \textbf{HI} segment}\\
n_c &= \text{number of clusters}\\
\mathbf{Y} = \{\mathbf{y}_i\in\mathbb{R}^{n}\}_{i=0}^{n_s-1} &= \text{the data to be clustered}\\
\mathbf{c} = \{\mathbf{c_j}\in\mathbb{R}^{n}\}_{j=0}^{n_c-1} &= \text{the cluster centers}\\
m &= \text{fuzziness coefficient, } m > 1
\end{align*}

The FCM algorithm minimizes the generalized least square error 
\begin{equation}
\label{eq:eq001}
J_m(\mathbf{\delta},\mathbf{c};\mathbf{Y}) = \sum_{i=0}^{n_s-1}\sum_{j=0}^{n_c-1}(\delta_{ij})^m(\mathbf{y_i}-\mathbf{c_j})^T\cdot A\cdot(\mathbf{y_i}-\mathbf{c_j})
\end{equation}
where $\delta_{ij}$ is the likelihood of $\mathbf{y_i}$ belonging to cluster $\mathbf{c_j}$, $A\in \mathbb{R}^{n\times n}$ is a positive definite weight matrix determined by the type of cluster. In this work we consider hyperspherical cluster with $A = I$, the identity matrix. Under this condition, the FCM algorithm minimizes the least square Euclidean norm of Equation \ref{eq:eq001}. As discussed in \citep{bezdek1984fcm}, at the local minimum of $J_m(\mathbf{\delta},\mathbf{c};\mathbf{Y})$,
\begin{align}
\label{eq:eq002}
\mathbf{c_j} &= \frac{\sum_{i=0}^{n_s-1}(\delta_{ij})^m\cdot \mathbf{y_i}}{\sum_{i=0}^{n_s-1}(\delta_{ij})^m} &0\leq j \leq n_c-1\nonumber\\
\delta_{ij} &= \Bigg(\sum_{k=0}^{n_c-1}\Big(\frac{(\mathbf{y_i}-\mathbf{c_j})^T\cdot(\mathbf{y_i}-\mathbf{c_j})}{(\mathbf{y_i}-\mathbf{c_k})^T\cdot(\mathbf{y_i}-\mathbf{c_k})}\Big)^{\frac{1}{m-1}}\Bigg)^{-1} & 0\leq i\leq n_s-1, 0\leq j\leq n_c-1
\end{align}
Due to the cyclic dependency of cluster centers on the likelihood and vice-versa, Equation \ref{eq:eq002} is solved iteratively for a fixed number of iterations \citep{cannon1986efficient}.

\subsubsection{Dimensionality Reduction And Cluster Center Selection}
Unlike the native FCM, which involves selecting the cluster centers for a fixed dimension, our approach involves selecting (1) the optimum set of neurons (dimensionality reduction), together with (2) the cluster centers\footnote{$n\in\mathbb{Z}$, while the cluster centers $\mathbf{c_j}\in\mathbb{R}^n$; solution to Equation \ref{eq:eq001} is therefore of mixed-integer type.}. The problem of finding the optimum dimension is also referred to as feature selection \citep{keogh2011curse}. Standard feature selection techniques such as sequential forward (backward) selection, SFS (SBS) \citep{whitney1971direct,marill1963effectiveness} are often associated with high computation cost and are prone to being stuck at local minimum. We use particle swarm optimization (PSO) \citep{eberhart1995new}, an evolutionary computing technique inspired by social behaviors such as bird flocking and fish schooling. Evolutionary computing techniques are efficient in avoiding to stuck at local optima. Additionally, PSO is computationally less expensive with faster convergence compared to its counterparts such as genetic algorithm (GA) or simulated annealing (SA), which makes PSO the ideal candidate for resource and power constrained wearable devices.

\begin{figure}[t]
	\hfill
	\begin{center}
		\includegraphics[scale=0.5]{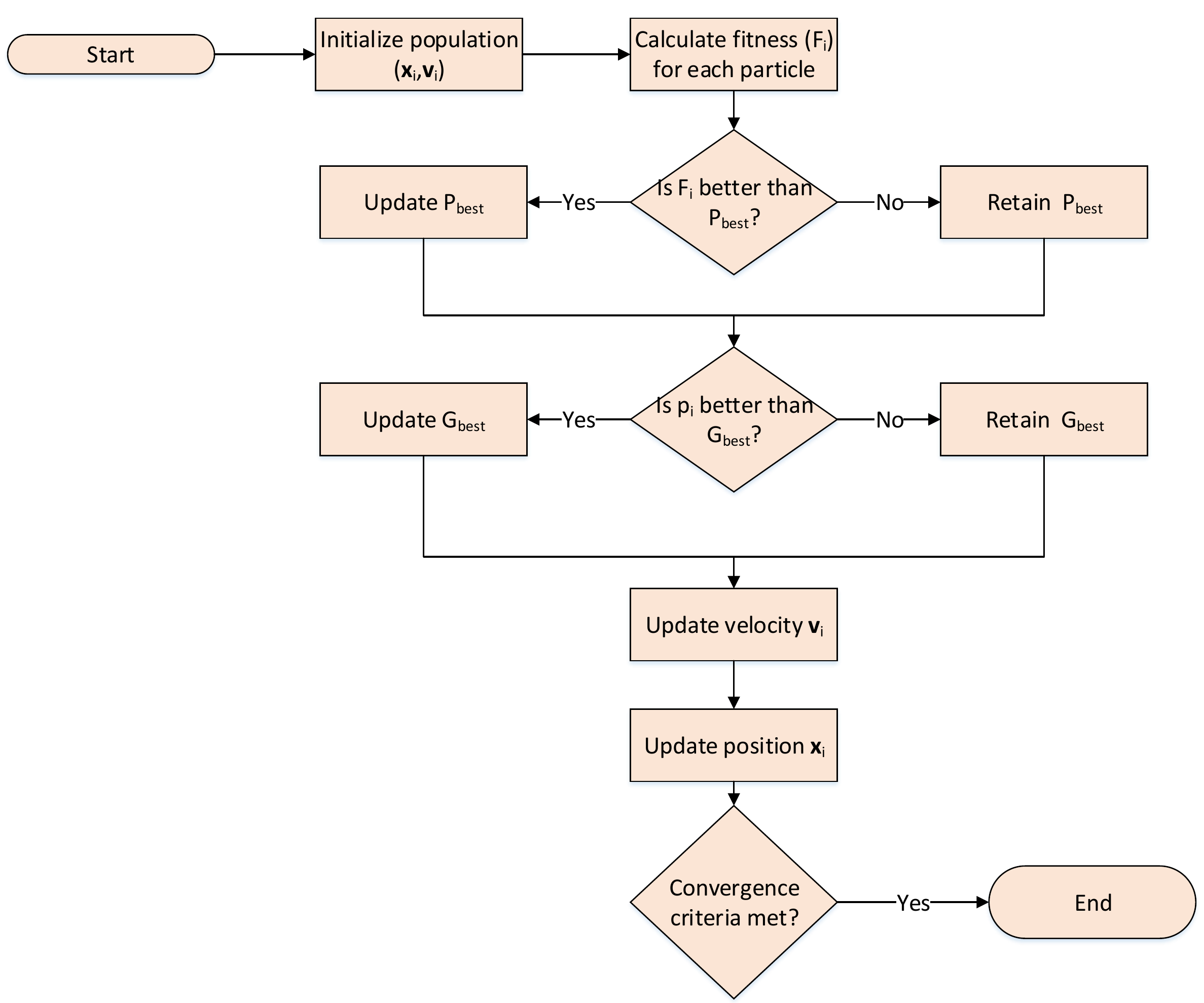}
	\end{center}
	\caption{Particle swarm optimization to find an optimum solution.}
	\label{fig:pso}
\end{figure}

In general, PSO finds the optimum solution to a fitness function $F$. Each solution is represented as a particle in the swarm. Each particle has a velocity with which it moves in the search space to find the optimum solution. During the movement, a particle updates its position and velocity according to its own experience (closeness to the optimum) and also experience of its neighbors. We introduce the following new notations related to PSO.
\begin{align*}
D &= \text{dimensions of the search space}\\
n_p &= \text{number of particles in the swarm, i.e., the swarm size}\\
\mathbf{\Theta} = \{\mathbf{\theta}_l\in\mathbb{R}^{D}\}_{l=0}^{n_p-1} &= \text{positions of all the particles in the swarm}\\
\mathbf{V} = \{\mathbf{v}_l\in\mathbb{R}^{D}\}_{l=0}^{n_p-1} &= \text{velocity of all the particles in the swarm}
\end{align*}

Position and velocity updates are performed according to the following equations.
\begin{align}
\label{eq:eq101}
\mathbf{\Theta}(t+1) = \mathbf{\Theta}(t) + \mathbf{V}(t+1)\nonumber\\
\mathbf{V}(t+1) = \mathbf{V}(t) + \varphi_1\cdot\Big(P_{\text{best}}-\mathbf{\Theta}(t)\Big) + \varphi_2\cdot\Big(G_{\text{best}}-\mathbf{\Theta}(t)\Big)
\end{align}
where $t$ is the iteration number, $\varphi_1,\varphi_2$ are constants and $P_{\text{best}}$ (and $G_{\text{best}}$) is the particles own (and neighbors) experience. Figure \ref{fig:pso} shows the iterative solution of PSO, where position and velocity are updated until a predefined convergence is achieved.

\noindent\underline{\emph{PSO for dimensionality reduction:}} We introduce $\mathbf{w} = \{w_d\}_{d=0}^{n-1}$, $w_d\in\{0,1\}$, a weighing factor, one for each output neuron. In order to transform real valued weights to binary values (for on/off decision of a neuron), two binarization functions are possible:
\begin{itemize}
	\item \emph{Deterministic Binarization:}
	\begin{equation*}
	\label{eq:102}
	\hat{w_d} = \texttt{sigmoid}(w_d-0.5)
	\end{equation*}
	\item \emph{Stochastic Binarization:}
	\begin{equation*}
	\label{eq:103}
	\hat{w_d} = \begin{cases}
	0 \text{~~with probability } \rho = \sigma(w_d)\\
	1 \text{~~with probability } 1 - \rho
	\end{cases}
	\end{equation*}
\end{itemize}

For simplicity we use deterministic binarization with $\mathbf{\hat{w}}$ representing the binarization operation performed on all elements of $\mathbf{w}$. Modified neuron spike response is
\begin{equation}
\label{eq:eq104}
\mathbf{\hat{y}_i} = \mathbf{y_i} \cdot \texttt{diag}(\mathbf{\hat{w}}) = \mathbf{y_i} \cdot \texttt{diag}\big(\texttt{sigmoid}(\mathbf{w}-0.5)\big)
\end{equation}
To use PSO for selecting the subset of output neurons, the weights $w_d$ are used as the dimensions of the particles in PSO. 

\noindent\underline{\emph{PSO for cluster centers:}} The number of clusters for the FCM algorithm is 2 (QRS and no-QRS) for the heart-rate estimation problem. This allows us to rewrite Equation \ref{eq:eq001} using Equation \ref{eq:eq104} as
\begin{align}
\label{eq:eq105}
&J_m(\mathbf{\delta},\mathbf{c};\mathbf{\hat{Y}}) = \sum_{i=0}^{n_s-1}\bigg((1-\delta_{i})^m(\mathbf{\hat{y}_i}-\mathbf{c_0})^T\cdot(\mathbf{\hat{y}_i}-\mathbf{c_0}) + (\delta_{i})^m(\mathbf{\hat{y}_i}-\mathbf{c_1})^T\cdot(\mathbf{\hat{y}_i}-\mathbf{c_1})\bigg)\nonumber\\
&\text{where }\delta_{i} = \delta_{i1} \text{~~~~and~~~~} (1-\delta_{i}) = \delta_{i0}
\end{align}

In order to use PSO to find the cluster centers there are two options:
\begin{itemize}
	\item \emph{to use each cluster center} as a PSO particle's dimension and use Figure \ref{fig:pso} to minimize the objective function $J_m(\mathbf{\delta},\mathbf{c};\mathbf{\hat{Y}})$, while using Equation \ref{eq:eq002} to determine the likelihood $\delta_{i}$ for every observation.
	\item \emph{to select each observation likelihood} as a PSO particle's dimension and use Figure \ref{fig:pso} to minimize the objective function $J_m(\mathbf{\delta},\mathbf{c};\mathbf{\hat{Y}})$, while using Equation \ref{eq:eq002} to determine the cluster centers $\mathbf{c_j}$.
\end{itemize}

Using likelihoods as PSO dimensions results in each particle in the swarm to be of $n_s\cdot n = 600n$ dimensions, causing the algorithm to converge at a slow rate to the optimum solution. We therefore use cluster centers as dimensions.

Combining, we have two cluster centers (each of dimension $n$) together with $n$ output neuron weights as combined PSO dimensions i.e., $D = 3n$. Under this assumption, the position of a swarm particle at $t^{\text{th}}$ iteration is 
\begin{align}
\label{eq:eq106}
&\theta_l(t) = \langle\theta_l^0,\theta_l^1,\cdots\theta_l^{3n-1}\rangle = \Big\langle \mathbf{w}(t),\mathbf{c_0(t)},\mathbf{c_1}(t)\Big\rangle\text{, where}\\
&\mathbf{w}(t) = \langle\theta_l^0,\theta_l^1,\cdots\theta_l^{n-1}\rangle = \text{ the weights selecting output neurons}\nonumber\\
&\mathbf{c_0}(t) = \langle\theta_l^n,\theta_l^{n+1},\cdots\theta_l^{2n-1}\rangle = \text{ coordinates of the center of cluster 0 (no QRS)}\nonumber\\
&\mathbf{c_1}(t) = \langle\theta_l^{2n},\theta_l^{2n+1},\cdots\theta_l^{3n-1}\rangle = \text{ coordinates of the center of cluster 1 (QRS)}\nonumber
\end{align}

For every iteration of the PSO algorithm, a particle's position is used to determine the set of neurons and the centers of the two clusters using Equation \ref{eq:eq106}. Subsequently, each observation $\mathbf{y}_i$ is modified according to Equation \ref{eq:eq104}, which is used to evaluate the fitness function using Equation \ref{eq:eq105}. When the PSO algorithm converges, Equation \ref{eq:eq002} is used to compute the probability $\delta_{i}$ of $\mathbf{y}_i$ belonging to cluster 1 (QRS cluster). The probability of $\mathbf{y}_i$ belonging to cluster 0 (no QRS cluster) is $1 - \delta_{i}$.
\subsubsection{Fuzzy c-Means Outcome to Discrete Heart-rate Probability Distribution}
In this section we use the probabilities $\delta_{i}$ to infer heart-rate. A naive solution is to use hard assignments as shown below.
\begin{equation}
\label{eq:eq201}
\text{heart-rate} = \sum_{i=0}^{n_s-1}\texttt{argmax}\Big((1-\delta_{i}),\delta_{i}\Big)
\end{equation}

In simple terms, this equation is interpreted as follows: if the probability of an observation belonging to the QRS cluster is higher than the probability of belonging to the no-QRS cluster, then the observation is interpreted as QRS and accounted for heart-rate estimation. In this work, we take a different approach. Rather than hard assignments as in Equation \ref{eq:eq201}, we compute the discrete heart-rate distribution from the probabilities computed using the FCM algorithm. In other words, we consider all observations (even the ones with low probability of association to the QRS cluster) to contribute towards heart-rate estimation. For this purpose, let $X_i$ be the random variable associated with the observation $\mathbf{y_i}$ with probability of success (i.e., belonging to QRS cluster) $p_i$. The $X_i$'s are independent random variables but are not identically distributed, as $p_i$'s can be different. This allows us to consider $X_i$'s as the outcome of Bernoulli's trials, with the sum of these variables (representing the heart-rate) forming the Poisson Binomial Distribution \citep{le1960approximation} of order $n_s$.
\begin{equation}
\label{eq:eq202}
X = \sum_{i=0}^{n_s-1} X_i
\end{equation}

We are interested in estimating the probability mass function (pmf) of $X$, which gives the discrete probability distribution of heart-rate. Let $\mathcal{S} = \{k_r~|~r\in 0,\cdots h-1\}$ be the set of $h$ unique observations in the \textbf{HI} interval with $k_r\in 0,1,\cdots,n_s-1$. Then,
\begin{equation}
\label{eq:eq203}
Pr(X=h) = \sum_{\forall \mathcal{S}}\prod_{k_r\in \mathcal{S}}p_{k_r}\prod_{\substack{i=0 \\ i\notin \mathcal{S}}}^{n_s-1}(1-p_i) = \text{probability of heart-rate to be } h
\end{equation}
where the above sum is evaluated over all possible ways of selecting $h$ observations from $n_s$. The solution to Equation \ref{eq:eq203} involves equating all the permutations for heart-rate $h$, which is intractable for larger $n_s$ and $h$. {We use Discrete Fourier Transform (DFT) of the characteristic function to determine the probability mass function (PMF) of the distribution \citep{zhang2017algorithm,hong2013computing}. For notational brevity, let $H$ be the maximum heart-rate of humans. PMF of the distribution be represented as $\{\Lambda_0,\Lambda_1,\cdots,\Lambda_{H-1}\}$, where}
\begin{equation}
\label{eq:eq203a}
\Lambda_{j} =  Pr(X=j)
\end{equation}
{Characteristic function of the random variable $X$ is computed as}
\begin{equation}
\label{eq:eq003b}
\varphi_X(t) = \mathbf{E}[\exp(itX)] = \mathbf{E}[\exp(it\sum_{k=0}^{n_s-1} X_k)], \text{ where } i = \sqrt{-1}
\end{equation}
{The above equation can be re-written as}
\begin{equation}
\label{eq:eq003c}
\varphi_X(t) = \prod_{k=0}^{n_s-1} [(1-p_k)+p_k\exp(it)]
\end{equation}
{The characteristic function can also be written in terms of the PMF as}
\begin{equation}
\label{eq:eq003d}
\varphi_X(t) = \sum_{j=0}^{H-1} \Lambda_j \exp(itj)
\end{equation}
{Equating Equations \ref{eq:eq003c} \& \ref{eq:eq003d}},
\begin{equation}
\label{eq:eq003e}
\sum_{j=0}^{H-1} \Lambda_j \exp(itj) = \prod_{k=0}^{n_s-1} [(1-p_k)+p_k\exp(it)]
\end{equation}
{Expressing $t = \frac{2\pi n}{H}$, Equation \ref{eq:eq003e} can be rewritten as}
\begin{equation}
\label{eq:eq003f}
\sum_{j=0}^{H-1} \Lambda_j \exp(\frac{i2\pi n j}{H}) = \prod_{k=0}^{n_s-1} [(1-p_k)+p_k\exp(\frac{i2\pi n}{H})] = H\lambda_n
\end{equation}
{where}
\begin{equation}
\label{eq:eq003g}
\lambda_n = \frac{1}{H}\prod_{k=0}^{n_s-1} [(1-p_k)+p_k\exp(\frac{i2\pi n}{H})]
\end{equation}
{Observe the similarity of Equation \ref{eq:eq003f} to that of inverse discrete Fourier Transform (IDFT), allowing us to compute the PMFs $\{\Lambda_0,\Lambda_1,\cdots,\Lambda_{H-1}\}$ as the discrete Fourier Transform (DFT) of the characteristic function $\lambda_n$.}

\noindent {The expected heart-rate can be represented as} 
\begin{equation}
\label{eq:eq204}
\text{heart-rate} = \mathbf{E}(X) = \sum_{j=0}^{H-1} j\cdot \Lambda_j
\end{equation}


\begin{table}[t]
	\renewcommand{\arraystretch}{0.8}
	\caption{ECG databases used for evaluation of our proposed approach.}
	\label{tab:ecg}
	\begin{scriptsize}
		\begin{center}
			\begin{tabular}{|l|c|c|c|}\hline
				Database & Subjects & Duration (mins) & Sampling Rate (Hz)\\ \hline
				Internal (intdb) & 8 & $>$ 10 & 256\\
				MIT BIH Supraventricular Arrhythmia Database (svdb) & 14 & $>$ 30 & 128\\
				MIT-BIH Long Term Database (ltdb) & 7 & $>$ 60 & 128\\
				Long Term ST Database (ltstdb) & 31 & $>$ 60 & 250\\ \hline
			\end{tabular}
		\end{center}
	\end{scriptsize}
\end{table}

\section{Results and Analysis}
{We evaluated our approach on ECG recordings from 8 subjects collected during in-house clinical trials.} To demonstrate the applicability of our approach to wider subjects, three other databases are considered from open source Physiobank recordings \citep{goldberger2000physiobank}. Details of these databases are reported in Table~\ref{tab:ecg}. {It is to be noted that subjects from our internal database consists of healthy subjects. Additionally, subjects from svdb have cardiac arrhythmia, subjects from ltdb have normal sinus rhythms and subjects from ltstdb have myocardial ischemia. These databases are chosen to demonstrate the applicability of our approach to subjects with healthy cardiac rhythms and also to the ones with different types of cardiac irregularities.}

\subsection{In-house Clinical Data Collection Setup}
{In-house ECG data were collected as part of the variability study of ECG and PPG heart-rate intervals. These data were collected using Osram PPG module (test) and gtec (g.USBamp) biosignal amplifier and acquisition system (reference). All the recruited subjects have no history of cardiac disorders. The experimental setup consists of}
\begin{itemize}
	\item {Osram module (SFH7060) with three green and one each of red and infrared LEDs.}
	\item {Texas Instruments (TI) analog front end (TI AFE 4403 EVM; evaluation kit).}
	\item {gtec (g.USBamp) biosignal amplifier and acquisition system (CE-certified and FDA-listed medical device, safety class: II, conformity class: IIa).}
	\item {Workstation/laptop installed with gtec and TI data acquisition software.}
	\item {Osram PPG module mounted on a Velcro band for strapping to subject’s wrist during measurements.}
	\item {3 pre-gelled snap electrodes (ECG measurements).}
\end{itemize}

Data collection methods for the public databases are partially described in \cite{goldberger2000physiobank} for svdb and ltdb and in \cite{jager2003long} for ltstdb. 

\subsection{Spiking Neural Network Simulation Setup}
{Our simulation\footnote{{The simulation setup is used here for accelerating ECG simulations. Our final approach is envisioned to be running on a neuromorphic chip such as the CxQuad \cite{indiveri2015neuromorphic} for real-time ECG-based estimation.}} framework consists of}
\begin{itemize}
	\item {A Python module to convert ECG to spikes.}
	\item {The core CARLsim simulator (in C++) implementing the Liquid.}
	\item {A Python readout implementing unsupervised heart-rate estimation.}
\end{itemize}
{The simulation is carried on a system with the following configuration}
\begin{itemize}
	\item {CPU: Intel Core-i7 8 cores}
	\item {GPU: Nvidia GeForce GTX970}
	\item {Memory: 32GB}
	\item {OS: Ubuntu 14.04 (64 bits)}
\end{itemize}

Heart-rate estimated using our approach is compared with the golden heart-rate calculated using \cite{van2015345}, which uses computation intensive (and power hungry) deterministic QRS-detection techniques to compute the heart-rate. In the following subsections, we present our detailed evaluations.

\begin{figure}[t]
	\hfill
	\begin{center}
		\includegraphics[width=0.9\linewidth]{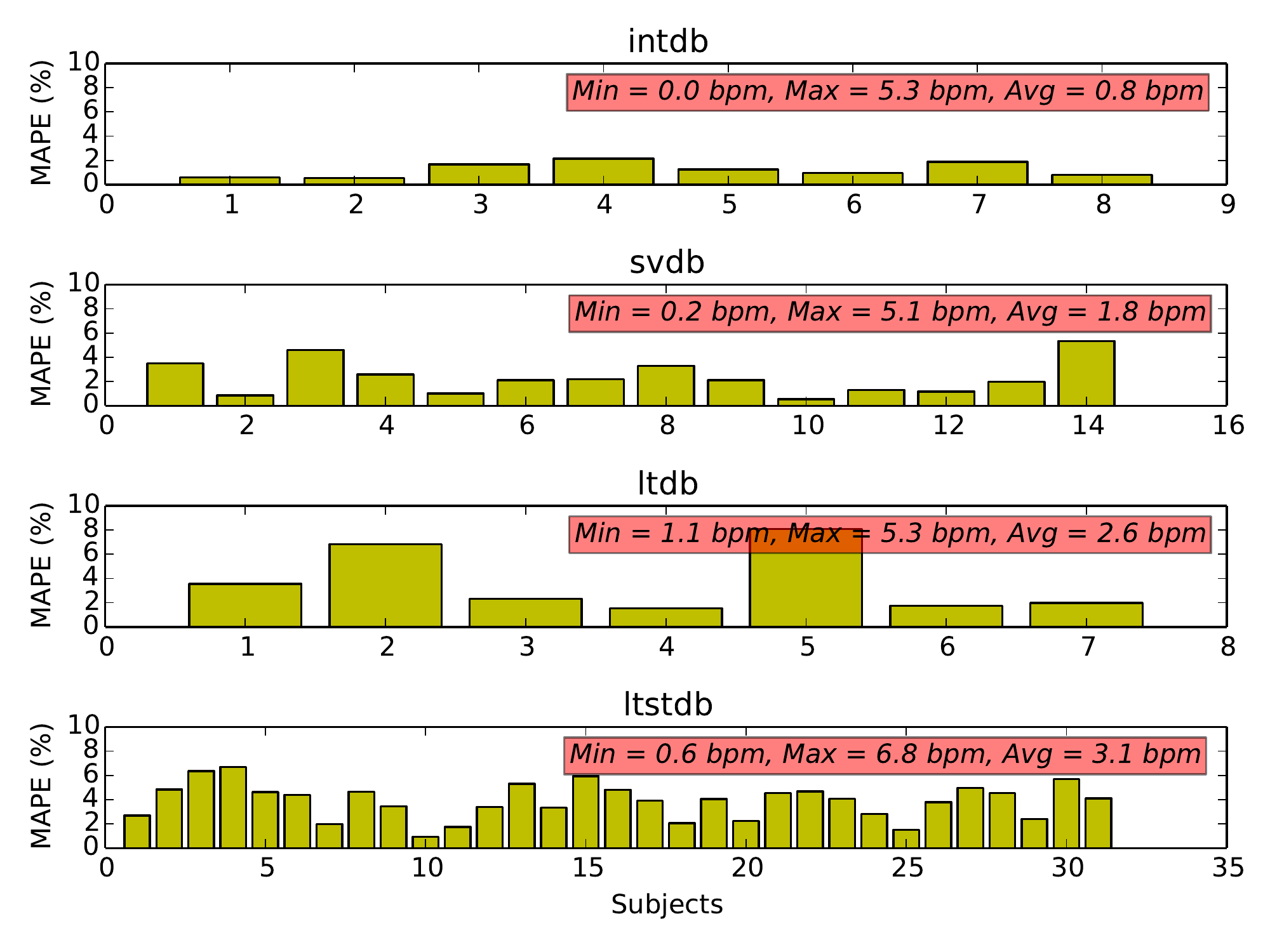}
	\end{center}
	\caption{Accuracy of our heart-rate estimation technique for all subjects of the four databases with respect to deterministic heart-rate. The accuracy is reported in terms of Mean Absolute Percentage Error (MAPE) for each subject (lower is better). Each figure also reports the statistics (minimum, maximum and average) of absolute heart-rate difference across all one minute segments for all patients in the database.}
	\label{fig:accuracy}
\end{figure}

\subsection{Heart-Rate Estimation Accuracy}
\label{sec:accuracy}
Figure \ref{fig:accuracy} plots the average heart-rate estimation accuracy using our approach for all subjects from four databases. The accuracy number for each subject is measured as Mean Average Percent Error (MAPE) calculated as
\begin{equation}
\text{MAPE} = \frac{1}{N}\sum_{i=0}^{N-1}\left(\frac{|a_i - p_i|}{a_i}\right)\times 100
\end{equation} 
where $|\cdot|$ is the L1 norm or the absolute difference between the actual heart-rate ($a_i$) and estimated heart-rate ($p_i$) and $N$ is the number of one-minute segments in the given ECG duration. As can be seen from the figure, the MAPE using our approach is less than 10\% across all subjects in all databases. For our internal database, the MAPE varies between 0.5\% and 2.1\%, with an average of 1.2\%. In terms of absolute values, these subjects have heart-rate differences between 0 and 5.3 beats per minute (bpm), with an average difference of 0.8 bpm.

For svdb database, the MAPE varies between 0.53\% and 5.3\%, with an average of 2.32\%. The heart-rate difference of the subjects are in the range of 0.2 bpm to 5.1 bpm, an average of 1.8 bpm. For ltdb, the MAPE varies between 1.53\% and 8.07\% with an average of 3.54\%. Heart-rate difference of the subjects in this database are between 1.1 bpm and 5.3 bpm, an average 2.6 bpm. Finally for the ltstdb, the MAPE varies between 0.92\% and 6.7\% with an average of 3.88\%. The actual heart-rate difference varies between 0.6 bpm and 6.8 bpm, with an average difference of 3.1 bpm.

\subsection{Intuition Behind Mis-prediction and Possible Solution}
\begin{figure}[t]
	\hfill
	\begin{center}
		\includegraphics[width=0.9\linewidth]{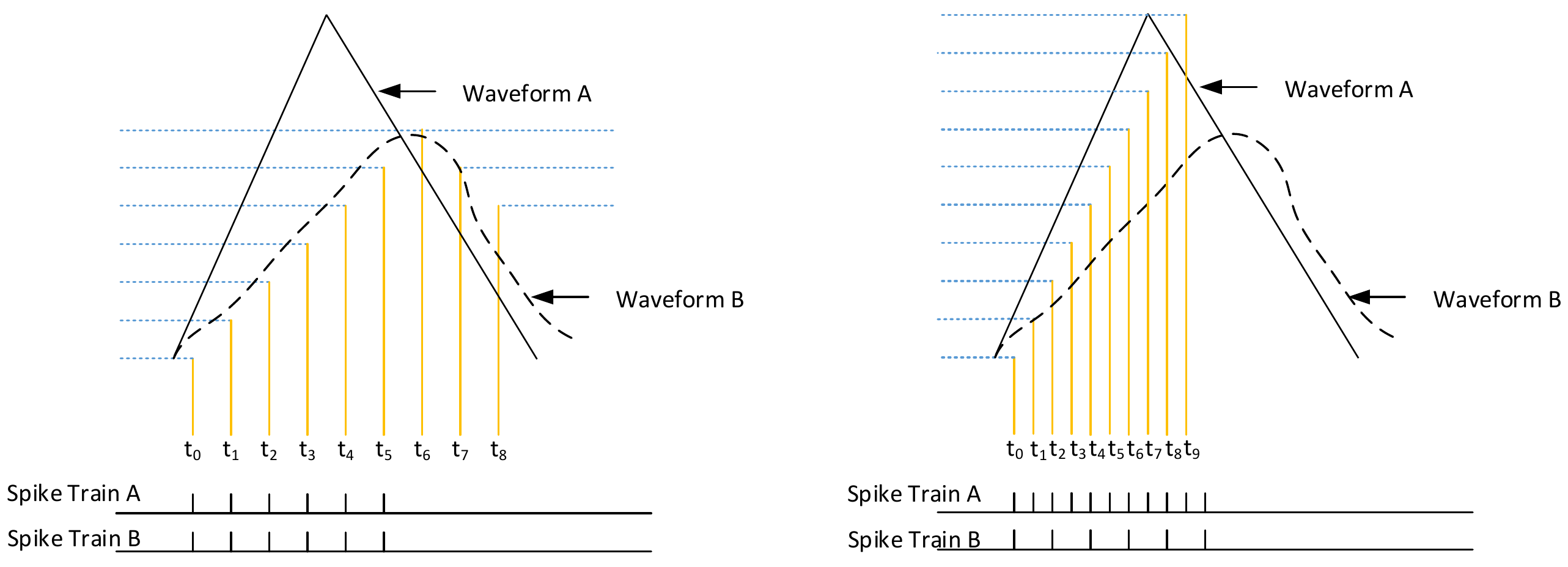}
	\end{center}
	\caption{Spike train generated for two waveforms (left). As seen from this figure, the two waveforms are indistinguishable when we compare their spike trains. A possible solution is an adaptive spike generation (right).}
	\label{fig:aliasing}
\end{figure}

{As discussed in Section \ref{sec:spike_encoder}, we use a fixed value for the threshold gap \texttt{Delta} and the timer interval. While this fixed parameter spike encoding scheme works well for identifying most QRS complexes (high accuracy as seen in the previous subsection), there are some scenarios where our approach leads to mis-prediction (mis-detection or false detection, causing false negative or false positive, respectively). In this subsection, we discuss this and propose a simple solution to improve accuracy. Figure \ref{fig:aliasing}(a) shows two waveforms (A and B). In this figure $t_i$ are the fixed time instances when these waveforms are compared with the thresholds. Spikes generated from the two waveforms are shown at the bottom of this figure. As seen from this figure, spike trains of the two waveforms are similar. These causes aliasing effect. In other words, the spiking neural network will learn to treat the two waveforms similarly (both as QRS or both as not QRS). This results in mis-prediction in some cases as seen from Figure~\ref{fig:accuracy}.}

{One technique to overcome this situation is to adapt the timer interval in Figure~\ref{fig:circuit} in response to the input waveform. This is shown in figure \ref{fig:aliasing}(b), where the timer interval for waveform A is reduced to better track the fast rising waveform A. Spike trains generated from the waveform are shown in the bottom of this figure. It is to be noted that the timer interval for waveform B is unaltered (same as in Figure \ref{fig:aliasing})(a). This results in the spike train shown at the bottom for this waveform. It can be seen quite clearly that spike trains for waveforms A and B are now different. This will allow our spiking neural network to better distinguish the two spike trains, further improving accuracy. One approach to adapt timer intervals is by setting its clock frequency to be proportional to the highest Fourier component extracted from the waveform. This will be addressed in future.}


\begin{figure}
	\centering     
	\subfigure[Heart-rate difference for a subject in intdb.]{\label{fig:imec_histogram}	\includegraphics[width=2.3in]{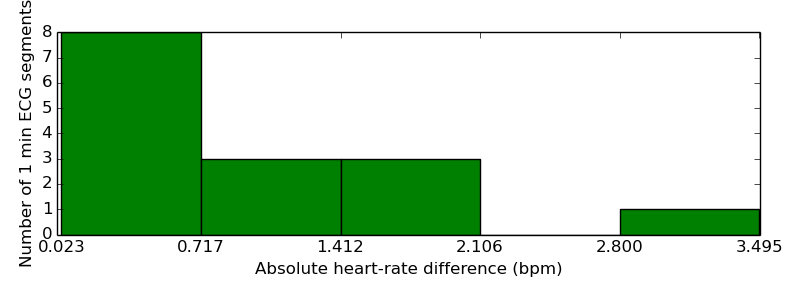}}
	\subfigure[Heart-rate difference for a subject in svdb.]{\label{fig:svdb_histogram}	\includegraphics[width=2.3in]{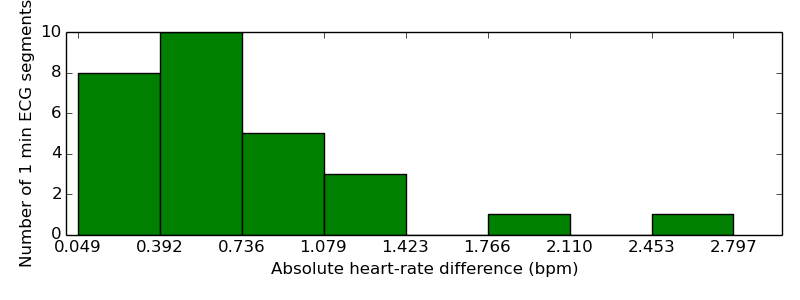}}
	\subfigure[Heart-rate difference for a subject in ltdb.]{\label{fig:ltdb_histogram}	\includegraphics[width=2.3in]{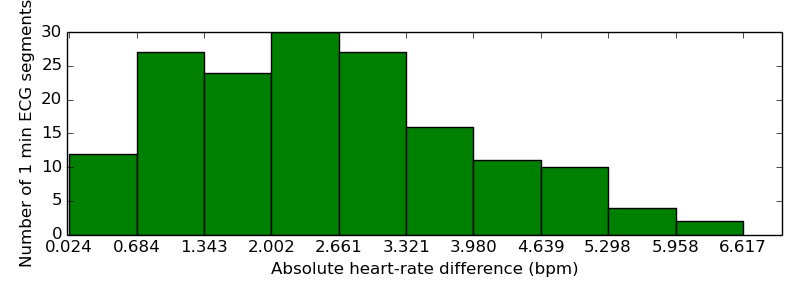}}
	\subfigure[Heart-rate difference for a sunject in ltstdb]{\label{fig:ltstdb_histogram}	\includegraphics[width=2.3in]{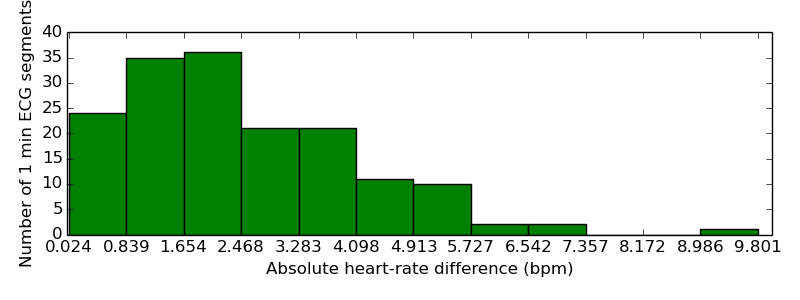}}
	\caption{Heart-rate difference histogram for representative subjects in four databases.}
	\label{fig:hrd_histogram}
\end{figure}

\subsection{Histogram of Heart-rate Difference}
Figure \ref{fig:hrd_histogram} plots the difference between the estimated and the actual heart-rate for representative subjects from the four databases. Each sub-figure plots the absolute heart-rate difference in bpm (on the x-axis) and the distribution i.e., the number of one minute segments on the y-axis. As seen for the subject from intdb in Figure \ref{fig:imec_histogram}, out of 13 one minute ECG segments, 8 have heart-rate difference of 0.7 bpm or lower, another 4 segments have heart-rate difference between 0.7 bpm and 2.1 bpm. Finally, there is one segment with heart-rate difference between 2.8 and 3.5 bpm. Results are similar for other subjects of the same database. Similarly, for the subject from svdb in Figure \ref{fig:svdb_histogram}, 26 out of 28 one minute segments have heart-rate difference of 1.4 bpm or lower, the two remaining segments have heart-rate difference of 2.3 bpm and 2.8 bpm, respectively. Results are similar for the two representative subjects from ltdb and ltstdb, which are shown in Figures \ref{fig:ltdb_histogram} and \ref{fig:ltstdb_histogram}, respectively. 


\subsection{Comparison with other learning-based heart-rate estimation techniques}
\label{sec:sota}
In this subsection, we compare other artificial intelligence techniques applied to the heart-rate estimation problem. Specifically, we use support vector machine (SVM) implemented with Scikit Python toolbox \citep{pedregosa2011scikit} and convolutional neural network (ConvNet) implemented in tensorflow \citep{abadi2016tensorflow}, both of which are supervised learning. {These toolboxes are used to implement reduced versions of the approaches presented in \citep{magrans2016myocardial} and \citep{acharya2017automated}, respectively.} In contrast to these approaches, ours is an unsupervised approach. Summary of these results are presented in Table \ref{tab:sota}. In this table, Column 1 reports the method used for heart-rate estimation; Column 2 reports the underlying algorithm; Columns 3-6 report the average heart-rate estimation accuracy (in terms of MAPE with respect to using \cite{van2015345}) across all subjects in the four databases. It is to be noted that supervised approaches are sensitive to training set selection. To showcase this, we report four numbers each, for the two supervised approaches. The first number is the accuracy obtained when trained with ECG data from regular subjects (i.e., subjects with no cardiac irregularities). The second number is the accuracy when the supervised approaches are trained with subjects from svdb (with arrhythmia). The third number is the accuracy when trained with subjects from ltstdb (with myocardial ischemia). Finally, the fourth number reports the accuracy when trained with representative subjects with and without cardiac irregularities. {The training set is the same for the two supervised approaches, while test set is the same for all three approaches.} 

\begin{table}[t]
	\renewcommand{\arraystretch}{1.2}
	\caption{{Comparison with other state-of-the-art approaches in terms of Mean Absolute Percentage Error (MAPE) for heart-rate estimation (in percentage). The MAPE is computed with respect to the deterministic technique of \citep{van2015345}, which is computationally intensive and power hungry. Lower MAPE is better.}}
	\label{tab:sota}
	\begin{scriptsize}
		\begin{center}
			\begin{tabular}{|l|c|c|c|c|c|}\hline
				Methods & Approach & intdb & svdb & ltdb & ltstdb\\ \hline
				SVM \citep{magrans2016myocardial}& Supervised & 2.0\textbf{\big/}6.9\textbf{\big/}8.2\textbf{\big/}5.2 & 9.8\textbf{\big/}2.3\textbf{\big/}8.1\textbf{\big/}4.4 & 3.8\textbf{\big/}9.4\textbf{\big/}11.9\textbf{\big/}6.9 & 11.6\textbf{\big/}14.1\textbf{\big/}3.7\textbf{\big/}8.4\\
				ConvNet \citep{acharya2017automated} & Supervised & 2.5\textbf{\big/}5.5\textbf{\big/}10.0\textbf{\big/}5.7 & 11.1\textbf{\big/}2.1\textbf{\big/}10.1\textbf{\big/}4.6 & 3.6\textbf{\big/}8.9\textbf{\big/}8.9\textbf{\big/}3.9 & 10.3\textbf{\big/}21.1\textbf{\big/}3.7\textbf{\big/}7.4\\
				\hline
				Our Approach & Unsupervised & 1.2 & 2.32 & 3.54 & 3.88\\
				\hline
			\end{tabular}
		\end{center}
	\end{scriptsize}
\end{table}


{Following observations can be made from results presented in Table \ref{tab:sota}. SVM achieves better accuracy on ltdb and some subjects from intdb, when the classifier is trained with regular subjects (first numbers in row 2, columns 3 and 5). For svdb and ltstdb, the classifier performs poorly (MAPE of approximately 10\% and 12\% for these two databases, respectively as seen from first numbers in row 2, columns 4 and 6). On the other hand, when SVM is trained with subjects from arrhythmia database (svdb), accuracy for this database increases (2.3\%, second number in row 2, column 4). However, accuracies for the other databases are lower (second numbers in row 2, columns 3, 5, 6). Similarly, accuracy is better for ltstdb, when SVM is trained with subjects of this database (third number in row 2, column 6). In this case also the accuracies for other three databases are low (third numbers in row 2, columns 3, 4, 5). This trend is consistent for the convolutional neural network based classifier. From these trend its can be concluded that supervised training approaches need to be specialized for the particular cardiac condition, in order to achieve a good heart-rate estimation accuracy.}

{A second observation can be made from the above results. If the SVM is trained with representative subjects from all databases, average accuracy across these databases increases. However, many subjects in these databases have over 20\% error. This is also true for the ConvNet based approach. These results signify that a pre-trained model, even with specialized training can result in high error, when applied to unknown subjects. In contrast to these two approaches, our approach learns from the ECG data itself, eliminating the need for a general training set. Additionally, our approach can be applied to new subjects, while consistently providing high estimation accuracy.}



\subsection{Data Density and Energy Reduction}
\label{sec:energy}
Table \ref{tab:dce} reports the data density and energy reduction summary for subjects in the four databases. For each row item in the table, we report the ECG duration (in sec) in column 2, average spike firing rate (in Hz) in column 3, the data density (bits per spike) in column 4 and the energy consumption (in $\mu$J) in column 5. The average spike firing rate is the total number of spikes generated in the spiking neural network averaged over the ECG duration (column 2). Data density measures efficiency of the spike encoder and is defined as
\begin{equation}
\label{eq:eq701}
\text{Data density} = \frac{\text{ADC bits}}{\text{input spikes}}
\end{equation}
where input spikes is the number of spikes generated at the output of our spike encoder (and input to the spiking neural network) and ADC bits is the number of bits per ECG sample transmitted post analog to digital conversion (ADC) for QRS detection in a standard non-spiking manner. {Finally, column 5 reports the energy reduction achieved using our approach compared to the approach of \citep{deepu2016ecg}. The energy consumption of our approach is for a state-of-the-art spiking hardware -- CxQuad \citep{indiveri2015neuromorphic}, using estimation methodology similar to \citep{cao2015spiking}. For fairness of comparison, the energy consumption using \cite{deepu2016ecg} excludes the ADC energy, while the energy consumption using ours exclude the spike encoder energy.} The table reports result for each subject in the intdb (rows 2-9, identified by intdb.S0x in column 1). Average results of these subjects is reported next (row 10, identified by intdb.avg). Finally, average across the subjects of the three other databases (svdb, ltdb and ltstdb) are reported in the next three rows (rows 11-13, identified by svdb.avg, ltdb.avg and ltstdb.avg, respectively).

\begin{table}[t]
	\renewcommand{\arraystretch}{1.2}
	\caption{Data density and energy reduction summary}
	\label{tab:dce}
	\begin{scriptsize}
		\begin{center}
			\begin{tabular}{|l|c|c|c|c|}\hline
				Subjects & ECG duration (sec) & Avg. Spike Firing Rate (Hz) & Data density (bits/spike) & Energy Reduction\\ \hline
				intdb.S01 & 570  & 7.2 & 40.6 & 38.7x\\
				intdb.S02 & 940  & 7.8 & 35.5 & 35.6x\\ 
				intdb.S03 & 940  & 8.5 & 35.5 & 32.7x\\
				intdb.S04 & 1060 & 5.7 & 63.8 & 48.6x\\
				intdb.S05 & 1060 & 7.8 & 33.8 & 35.6x\\
				intdb.S06 & 960  & 7.8 & 56.3 & 36.6x\\
				intdb.S07 & 720  & 8.9 & 39.5 & 31.2x\\
				intdb.S08 & 980  & 8.9 & 44.5 & 31.3x\\ 
				\hline
				intdb.avg & 896  & 7.8 & 43.7 & 36.2x\\
				\hline
				svdb.avg & 1790  & 7.6 & 52.6 & 37.1x\\
				\hline
				ltdb.avg & 9900  & 7.8 & 40.7 & 35.9x\\
				\hline
				ltstdb.avg & 9900  & 7.7 & 57.6 & 36.5x\\
				\hline
			\end{tabular}
		\end{center}
	\end{scriptsize}
\end{table}

As seen from the table, the average data density (averaged over all subjects) for our internal database (intdb) is 43.7. This result can be interpreted as follows: on average for every 43.7 bits of raw ECG data transmitted from the sensor for heart-rate detection using standard QRS-detection techniques, our spike encoder transmits one spike for the same purpose. This compression saves energy and data-bandwidth.The higher the density, the higher the savings. For the three other databases i.e., svdb, ltdb and ltstdb, our approach achieves data density of 52.6 (12.9 -- 113.3 for all subjects), 40.7 (24.4 -- 56.5 for all subjects) and 57.6 (21.7 -- 108.5 for all subjects), respectively.

{The average energy reduction using our heart-rate estimation approach (averaged over all subjects) for our internal database is 36.2 lower than \citep{deepu2016ecg}, signifying the importance of our approach for power constrained wearable devices. For the three other databases, the average energy savings of the subjects are 37.1x (32.9x -- 49.5x), 35.9x (33.7x -- 38.1x) and 36.5x (29.7x -- 46.6x), respectively. These results suggest that our approach can be very well integrated in future wearable devices, providing significant battery life and improving user experience.} 

\begin{table}[t]
	\renewcommand{\arraystretch}{1.2}
	\caption{QRS detection summary}
	\label{tab:qrs}
	\begin{scriptsize}
		\begin{center}
			\begin{tabular}{|l|c|c|c|c|c|c|}\hline
				\multirow{2}{*}{Subjects} & \multirow{2}{*}{Accuracy} & \multirow{2}{*}{False Positive} & \multirow{2}{*}{False Negative} & \multicolumn{3}{|c|}{ECG Detection Offset}\\\cline{5-7}
				&  &   &  & 0ms - 50ms & 50ms - 100ms & 100ms - 200ms\\ \hline
				intdb.S01 & 100\% (414) & 2.4\% & 0\% & 100\% & 0\% & 0\%\\
				intdb.S02 & 100\% (842) & 1.9\% & 0\% & 100\% & 0\% & 0\%\\
				intdb.S03 & 99.38\% (971) & 0.6\% & 0.6\% & 57.3\% & 39.3\% & 3.4\%\\
				intdb.S04 & 100\% (702) & 3.2\% & 0\% & 100\% & 0\% & 0\%\\
				intdb.S05 & 99.7\% (949) & 2.5\% & 0.3\% & 84.2\% & 12.3\% & 3.3\%\\
				intdb.S06 & 99.7\% (934) & 1.39\% & 0.3\% & 82.7\% & 17.3\% & 0\%\\
				intdb.S07 & 99.0\% (907) & 1.0\% & 1.0\% & 49.7\% & 39.7\% & 10.6\%\\
				intdb.S08 & 100\% (1169) & 1.19\% & 0\% & 100\% & 0\% & 0\%\\
				\hline
			\end{tabular}
		\end{center}
	\end{scriptsize}
\end{table}

\begin{figure}[t]
	\hfill
	\begin{center}
		\includegraphics[width=0.99\linewidth]{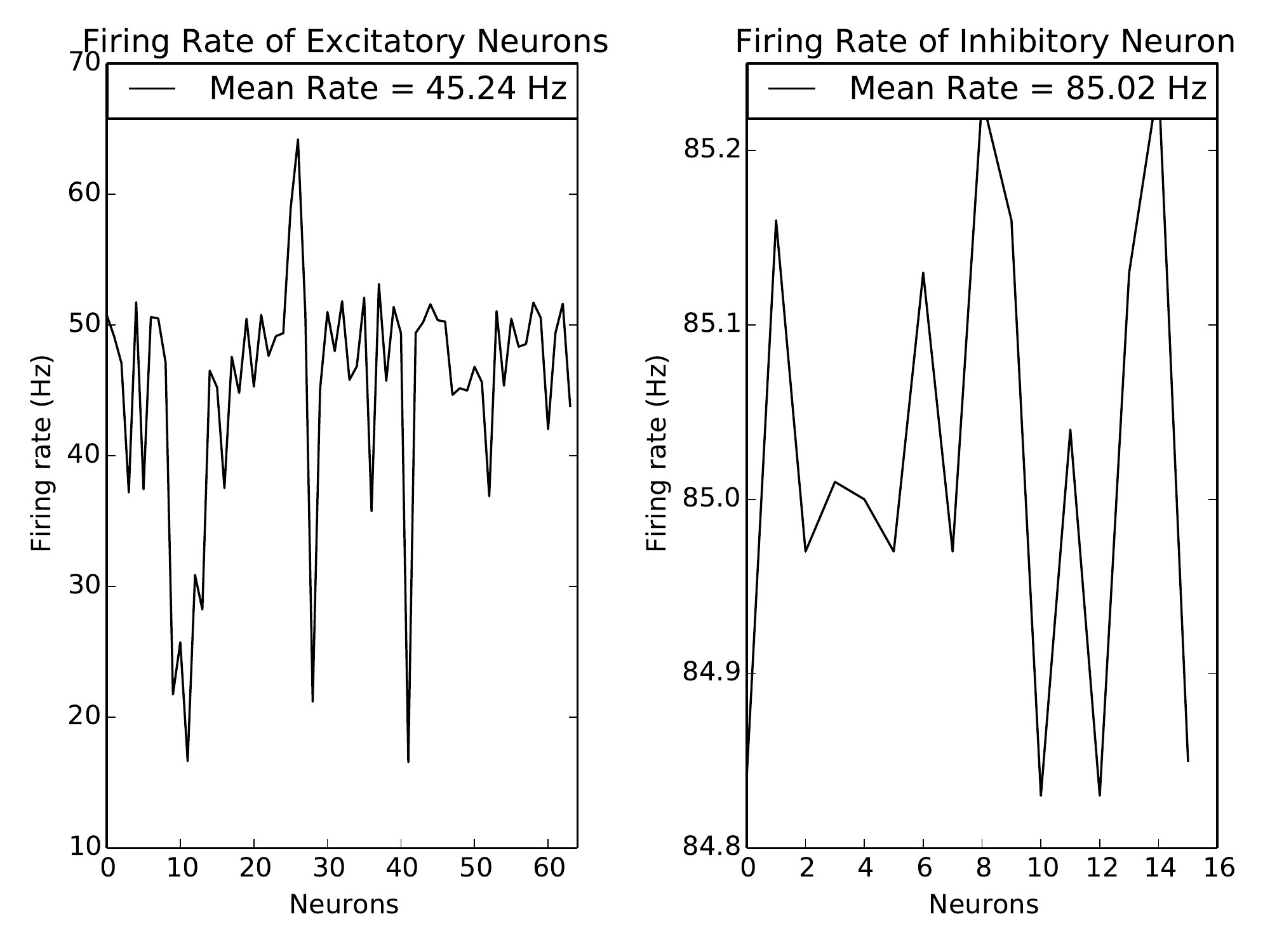}
	\end{center}
	\caption{Firing rate of the excitatory and inhibitory neurons.}
	\label{fig:fr}
\end{figure}

\subsection{Extension of our approach to other use-cases: QRS Detection Case Study}
\label{sec:qrs}
\newchanged{Our proposed approach can be easily extended to other use-cases by designing a new readout using the same Liquid. One most common use-case is the QRS detection, which is commonly used to detect clinically significant heart irregularities such as arrhythmia. In this section we show how our approach can be instantiated for QRS detection.} In this case, the probability distribution at the output of the FCM algorithm can be inferred as QRS based on hard assignments (similar to Equation \ref{eq:eq201}). In other words, if the probability of an observation being part of QRS cluster is higher than the probability of it to be part of no-QRS cluster, the observation is labeled as QRS. This can be implemented as a separate readout using the same Liquid. Table \ref{tab:qrs} reports QRS summary for the eight subjects in the intdb. In column 2 we report the detection accuracy in percentage for the actual number of QRS reported in parenthesis. Column 3 and 4 report respectively, the false positives and false negatives as percentage of the actual number of QRS. Column 5-7 report the offset between the detected QRS position and the position of actual QRS. Column 5 reports the percentage of the detected QRS with a maximum offset of 50ms, Column 6 that between 50ms to 100ms and Column 7 with offset between 100ms and 200ms. As can be seen from the table, our approach results in QRS detection accuracy of over 99\% for all subjects in the database. Additionally, the false positives and negatives are less 4\% and 1\%, respectively. Finally, over 90\% of the detected QRS have offset less than 100ms. It is to be noted that results using the three other databases (svdb, ltdb and ltstdb) are similar to that obtained for the intdb. These results signify the relevance of our approach when applied to QRS detection. {Our ongoing work is to investigate other use-cases, such as arrhythmia detection and ECG-based human authentication.}

\subsection{Spiking Neural Network Related Results}
\begin{figure}[t]
	\hfill
	\begin{center}
		\includegraphics[width=0.99\linewidth]{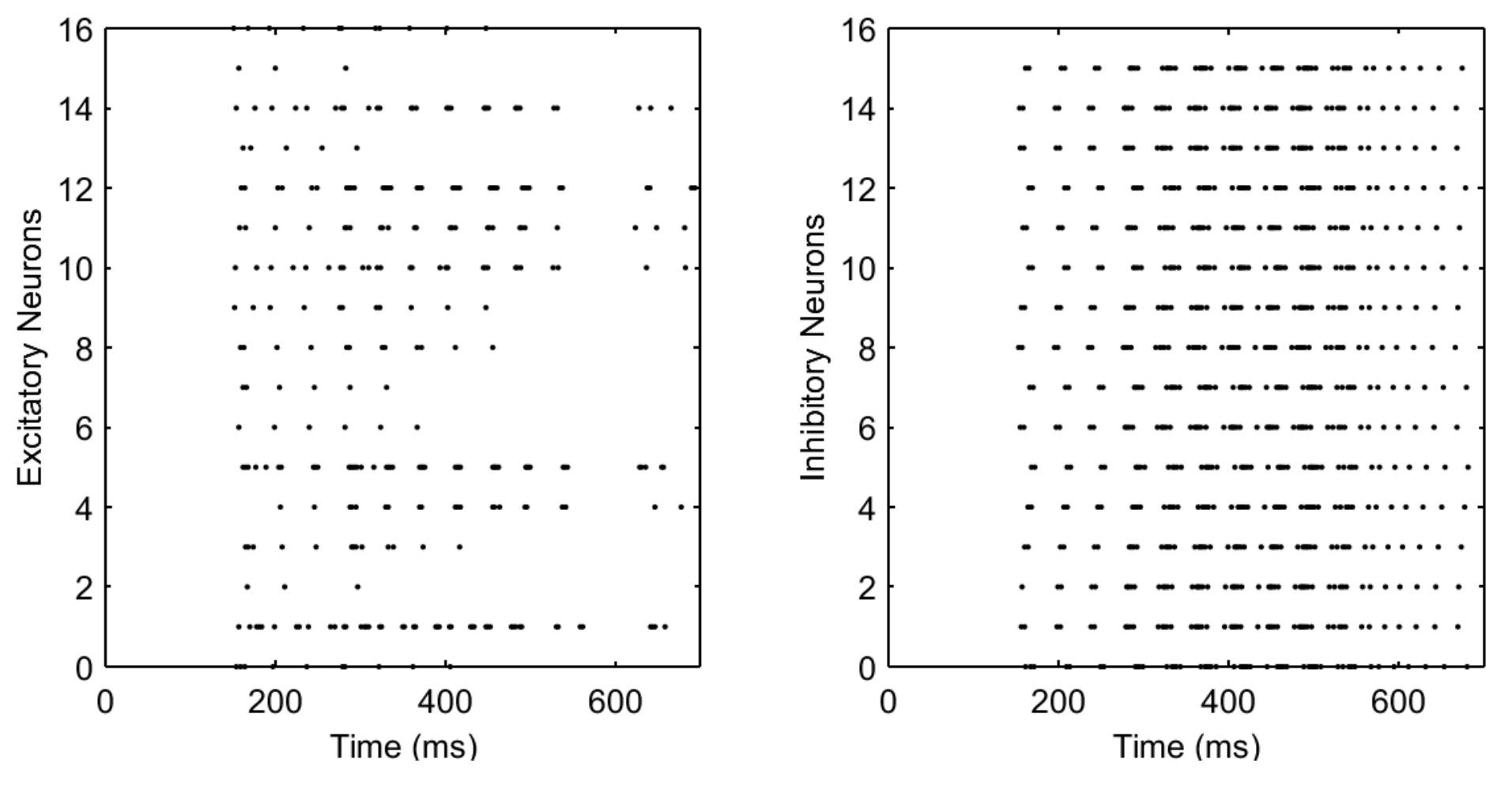}
	\end{center}
	\caption{Raster plot of the spike times for the excitatory and inhibitory neurons.}
	\label{fig:raster}
\end{figure}

\newchanged{This section provides results pertaining operation of the spiking neural network. Figure \ref{fig:fr} plots the firing rate of the excitatory and the inhibitory neurons of the spiking neural network Liquid. The average firing rate for the excitatory neurons is 45.24 Hz, while that for inhibitory neurons is 85.02 Hz. These results are for a specific subject selected randomly from the intdb. Results for subjects from other databases are similar. These firing rates are consistent with the firing rates reported in \cite{amit1997dynamics}. We performed design space exploration using the parameters mentioned in Table \ref{tab:param} to adjust the firing rates of the neurons. Finally, Figure \ref{fig:raster} plots the raster graph with spike times for 16 (out of 64) excitatory and 16 inhibitory neurons.}

\section{Conclusions and Outlook}
\newchanged{We presented an end-to-end approach for heart-rate estimation in wearables with embedded neuromorphic hardware. Starting from encoding an analog electrocardiogram (ECG) signal directly into spike train (temporal encoding), our approach uses a network of recurrently connected spiking neurons (the Liquid) to process these spike trains for inferring heart-rate. Finally, using a probabilistic inference (readout) unit at the output of the neural network, heart-rate is estimated in a completely unsupervised manner. The readout unit is designed using swarm intelligence to reduce dimensionality and select cluster centers of a Fuzzy c-Means Algorithm. Thorough evaluation using data from in-house clinical trials as well as with publicly available databases demonstrates the high accuracy of our approach together with significant data compression and low power consumption. Following are the useful insight from this work.}

\noindent\textbf{Relevance and Practicality:} \newchanged{Wearable electronic devices such as the activity tracker are gaining popularity because of their ability to monitor physical activity, sleep and other behaviors. A basic requirement of these devices is to track heart-rate accurately using limited energy for different cardiac conditions. Recent studies show that some of these devices suffer from poor accuracy and large power consumptions. Readers are referred to \cite{Evenson2015,sasaki2015validation} for a review of different activity trackers. Another limitation of these devices is that these devices offer limited flexibility to implement complicated clinically significant use-cases, such as detecting arrhythmia. This is due to small form factor of these devices coupled with limited flexibility and energy budgets. We have demonstrated that the end-to-end solution has a low energy footprint (see Section \ref{sec:energy}) together with high accuracy (see Section \ref{sec:accuracy}), using a limited number of neurons and synapses. This approach can be readily embedded in future wearable devices with strict energy and area budgets. The proposed LSM computational model offers flexibility by allowing implementation of clinically significant use-cases (as readouts) from the spatio-temporal properties of ECG integrated inside a network of spiking neurons (Liquid). We have demonstrated this using the QRS detection use-case in Section \ref{sec:qrs}. In future, we will investigate arrhythmia detection. Additionally, the unsupervised readout is conducive to personalized healthcare, by allowing learning from subjects directly, without requiring costly data annotations to train the network. This allows future wearables to be used seamlessly for subjects with and without cardiac irregularities.}

\noindent\textbf{Novelty:} \newchanged{Existing machine learning based ECG processing (QRS detection or heart-rate estimation) are primarily implemented using supervised learning. These techniques require good training sets to achieve acceptable accuracy as we demonstrated in Section \ref{sec:sota}. In addition to this, classical supervised approaches cannot be easily generalized to different cardiac irregularities. The power consumption is high due to the requirement of transmitting digitized bits between sensor and devices. To address these limitations, our approach presents three novel contributions: (1) a technique to encode spikes from ECG directly, without requiring to digitize the analog ECG signal and thus achieving over 40x reduction in data density (Section \ref{sec:energy}); (2) a novel learning rule for spiking neural network in a LSM computation model, which can be efficiently implemented on state-of-the-art neuromorphic hardware, with over 30x improvement in energy consumption (Section \ref{sec:energy}) over existing approaches; and (3) an unsupervised readout for heart-rate estimation case-study, where Liquid states are selected intelligently using Particle Swarm Optimization to improve Fuzzy c-Means clustering (Section \ref{sec:readout}). Unsupervised learning allows generalization to different cardiac conditions, eliminating the need for data annotation. Additionally, the approach can be readily deployed to subjects with rare cardiac conditions, where ECG data is not always available to train a supervised classifier. An example situation is heart-rate observation following Transcatheter Aortic Valve Implantation (TAVI) \cite{tamburino2011incidence}. TAVI subjects are currently monitored in lab facilities for a longer period after the implantation procedure. Our approach can be readily deployed for these subjects, facilitating early rehabilitation and remote monitoring.}

\noindent\textbf{Advancing neuromorphic computing: }\newchanged{Neuromorphic computing has matured significantly in recent years. This growth is fueled by innovations in materials \cite{kuzum2013synaptic,jo2010nanoscale,amit1994learning}, circuits \cite{pfeil2013six,snider2011synapses,chicca2014neuromorphic,walter2015neuromorphic}, algorithms \cite{beyeler2013categorization,shrestha2017robust,weng2013modulation} and applications \cite{CorradiTBME,verstraeten2005isolated,carneiro2013event}. Our work falls in the interface between neuromorphic applications and algorithms. Towards this end,  Table \ref{tab:nmc} reports some practical applications for neuromorphic computing based on LSM computation model. The table also indicates the contribution of this work. As can be seen, our proposed application contributes towards a real-life benchmark against which future neuromorphic architectures and algorithms can be evaluated.}


\begin{table}[t]
	\renewcommand{\arraystretch}{1.2}
	\caption{Some example applications for neuromorphic computing with LSM-based computation model.}
	\label{tab:nmc}
	\begin{scriptsize}
		\begin{center}
			\begin{tabular}{|l|c|c|c|}\hline
				Approach & Application Domain & Information Coding & Algorithm \\ \hline
				Verstraeten \emph{et al.} \cite{verstraeten2005isolated} & Speech Recognition & Temporal & LSM with supervised readout\\
				Grzyb \emph{et al.} \cite{grzyb2009facial} & Expression Recognition & Spatial Coding & LSM with supervised readout\\
				Corradi \emph{et al.} \cite{CorradiTBME} & EEG-based BMI &  Temporal & LSM with supervised readout\\ \hline
				Our work & ECG-based heart-rate estimation & Temporal &  LSM with unsupervised readout\\ \hline
			\end{tabular}
		\end{center}
	\end{scriptsize}
\end{table}


\section*{Acknowledgments}
We would like to acknowledge our long-term collaborators Dr. Federico Corradi and Prof. Giacomo Indiveri from the Institute of Neuroinformatics at University of Zurich for the useful discussions related to this work. This work is supported in parts by EU-H2020 grant NeuRAM3 Cube (NEUral computing aRchitectures in Advanced Monolithic 3D-VLSI nano-technologies) and ITEA3 proposal PARTNER (Patient-care Advancement with Responsive Technologies aNd Engagement togetheR).

\section*{References}

\bibliography{ECGsnn}

\end{document}